\documentclass[10pt,twocolumn,letterpaper]{article}

\usepackage{cvpr}              

\usepackage{graphicx}
\usepackage{amsmath}
\usepackage{amssymb}
\usepackage{booktabs}
\usepackage{algorithm}
\usepackage{algorithmic}
\usepackage{multirow}

\usepackage[pagebackref,breaklinks,colorlinks]{hyperref}

\usepackage[capitalize]{cleveref}
\crefname{section}{Sec.}{Secs.}
\Crefname{section}{Section}{Sections}
\Crefname{table}{Table}{Tables}
\crefname{table}{Tab.}{Tabs.}

\def\cvprPaperID{*****} 
\def\confName{CVPR}
\def\confYear{2022}

\begin{document}

\title{ARM3D: Attention-based relation module for indoor 3D object detection}

\author{Yuqing Lan, Yao Duan, Chenyi Liu, Chenyang Zhu, Yueshan Xiong \\
National University of Defense Technology\\
{\tt\small lanyuqingkd,duanyao16,liucheyi\_1013,zhuchenyang07@nudt.edu.cn,ysxiong@hotmail.com}\\

\and
Hui Huang\\
Shenzhen University\\
{\tt\small hhzhiyan@gmail.com}
\and
Kai Xu\\
National University of Defense Technology\\
{\tt\small kevin.kai.xu@gmail.com}

}
\maketitle

\begin{abstract}
  Relation context has been proved to be useful for many challenging vision tasks. In the field of 3D object detection, previous methods have been taking the advantage of context encoding, graph embedding, or explicit relation reasoning to extract relation context. However, there exists inevitably redundant relation context due to noisy or low-quality proposals. In fact, invalid relation context usually indicates underlying scene misunderstanding and ambiguity, which may, on the contrary, reduce the performance in complex scenes. Inspired by recent attention mechanism like Transformer, we propose a novel 3D attention-based relation module (ARM3D). It encompasses object-aware relation reasoning to extract pair-wise relation contexts among qualified proposals and an attention module to distribute attention weights towards different relation contexts. In this way, ARM3D can take full advantage of the useful relation context and filter those less relevant or even confusing contexts, which mitigates the ambiguity in detection. We have evaluated the effectiveness of ARM3D by plugging it into several state-of-the-art 3D object detectors and showing more accurate and robust detection results. Extensive experiments show the capability and generalization of ARM3D on 3D object detection. Our source code is available at
  \url{https://github.com/lanlan96/ARM3D}.
\end{abstract}

\section{Introduction}
\label{introduction}

\begin{figure}[!t]
  \centering
  \includegraphics[width=1.0\linewidth]{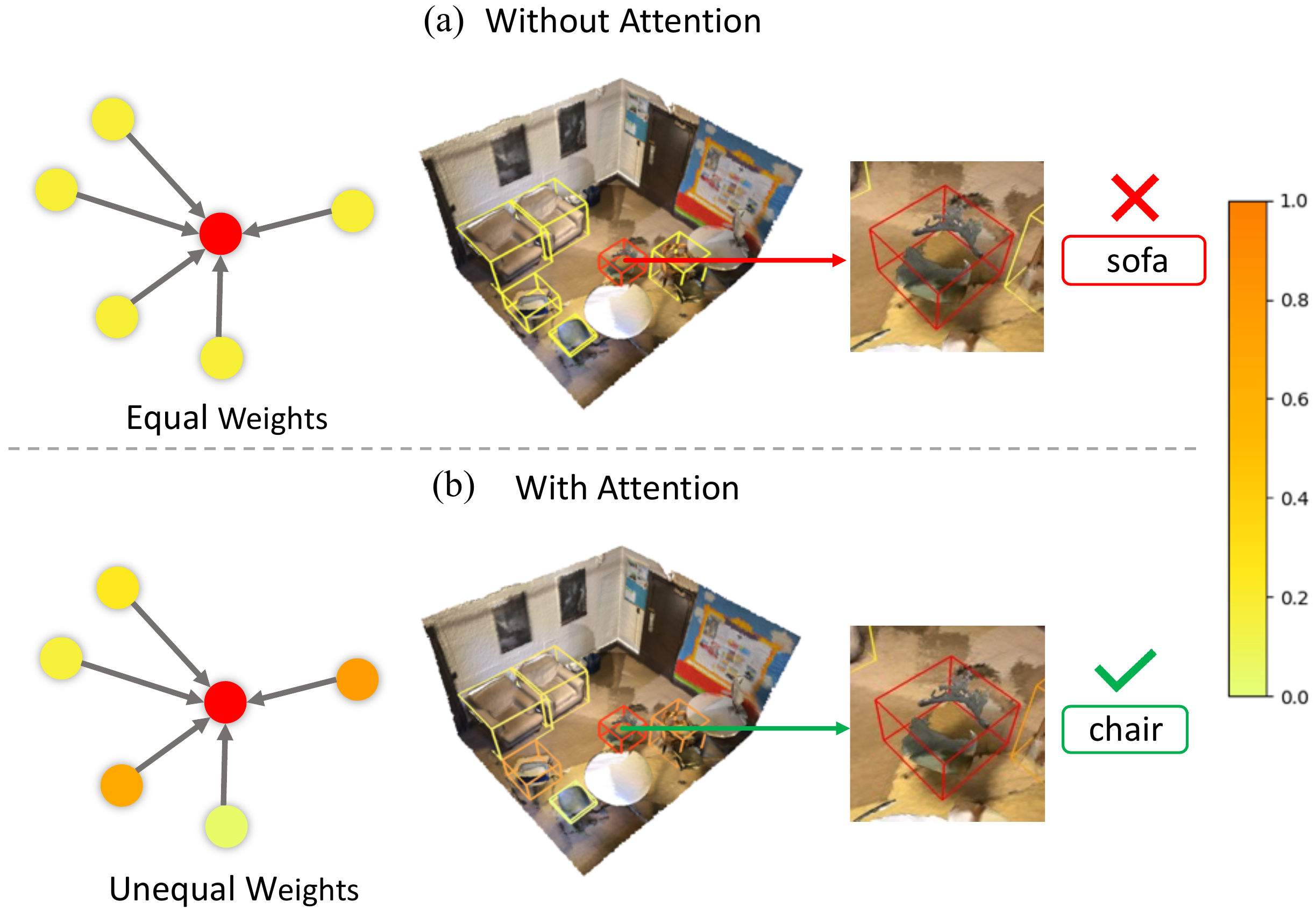}
  \caption{We propose an attention-based relation module (ARM3D) to reason about the most useful semantic relation contexts in 3D object detection. For example, all the objects with boxes in this figure are chairs represented as dots on the left. (a) A chair with the red box is hard to detect due to noise in point clouds and is mistakenly classified as a sofa using equal attention towards other objects. The upper left chairs in this scene have untypical structures, resulting in unclear semantic relations. (b) With unequal attention, this chair can pay more attention to the semantic relationships with objects having similar structures to filter the confusing context and thus can be classified correctly and robustly. Darker orange indicates greater attention.}
  \label{fig:teaser}
  \end{figure}

With the fast development of automatic and unmanned technology, 3D object detection has recently been brought to the fore. Nowadays, 3D object detection still remains challenging and plays an important role in 3D vision, including augmented reality, robot navigation, robot grasping, \emph{etc}. Most current 3D object detection methods focus on point clouds as a 3D representation: they are more readily available than before with the evolution of 3D scanning devices and reconstruction techniques. However, the orderless and unstructured nature of point clouds makes the detection in 3D more challenging than in 2D, as it is difficult to transfer widely used techniques for 2D object detection to 3D.

Recently, interests in point cloud have been on the rise to solve this challenge. With the boom in deep learning, more and more methods have been proposed to directly process 3D point clouds and use the extracted features for all kinds of 3D computer vision or graphics tasks~\cite{Charles2017PointNet,li2018pointcnn,qi2017pointnet++,wu2019pointconv}. Recent works~\cite{yi2019gspn,Qi_2019_ICCV,Xie_2020_CVPR,H3DNet,cheng2021brnet,lan20213drm} can effectively attain detected 3D objects in raw point clouds of indoor scenes. These methods mainly rely on the geometric features from deep backbones or contextual features from context encoding or relation reasoning.  

Context has been shown to be informative and useful in scene understanding~\cite{shi2016data,qi20173d,deepcontext} and is intuitively present in reality theoretically and practically. Nowadays, relation reasoning is playing an essential part in context modeling, which is applied to both 2D and 3D indoor object detection~\cite{hu2018relation,Xu_2019_CVPR,lan20213drm}. However, there are still two main unsolved challenges. On the one hand, most 3D detectors rely on proposals (object candidates) for classification and bounding box regression. Qualities of the proposals used in these methods are usually not satisfactory for extracting relation contexts, inevitably producing confusing or even improper contextual information. On the other hand, each proposal actually has its own specific needs for relation contexts from other proposals. Previous methods use equal weights for different relation contexts, which may ultimately result in more ambiguity or even misunderstanding (see Figure~\ref{fig:teaser}).

\begin{figure*}[h]

  \centering
    \includegraphics[width=1.0\linewidth]{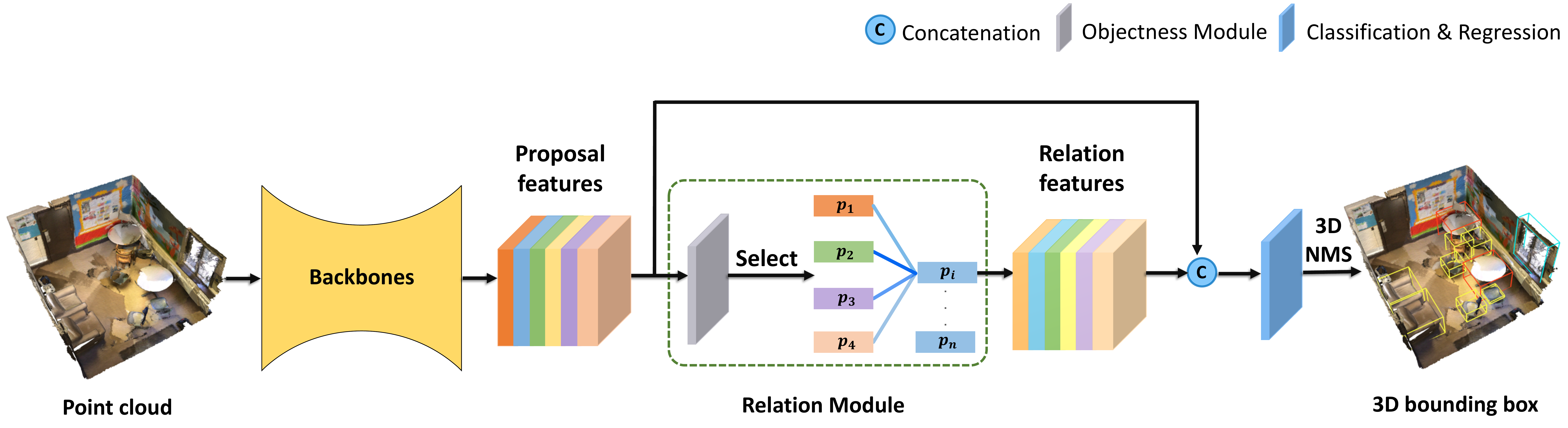}
  \caption{The 3D detection pipeline equipped with our ARM3D. With point cloud as input, the backbone networks of current proposal-based 3D detectors produce numerous proposals. These proposals are then sent into our attention-based relation module to extract the fine-grained relation features. These proposals are first selected according to their objectness, and each proposal is matched with several selected proposals to reason about their specific relation context. Darker blue means greater attention and higher weights.  The relation features are concatenated with the proposal features together. The combined features of different proposals are used by the detection heads to perform classification and regression. After 3D non-maximum suppression (NMS), the pipeline outputs the final detected bounding boxes.}
  \label{fig:framework}       
  \end{figure*}

In this paper, we propose an attention-based relation module for context modeling in 3D object detection to solve these two challenges. We argue that objects in indoor scenes are more or less relative to each other both semantically and spatially. As shown in Figure~\ref{fig:framework}, the core ideas of our novel method contain two parts which correspond to the two challenges respectively: Object-aware relation reasoning among different proposal pairs; An attention module based on Transformer to take full advantage of the most useful ones to extract contextual relation features. The first part includes a simple but quite useful objectness module to select proposals with high qualities. Available with selected proposals, we reason about both of the pair-wise semantic and spatial relations for different proposal pairs. As for the second part, we leverage an attention module based on Transformer to model the importance towards contexts from different proposal pairs for each selected proposal and thus reduce the effects of confusing contexts. In this way, we can not only enhance understanding and mitigate the ambiguity towards various objects in manifold indoor scenes but also avoid being affected by confusing or even useless context information together with the useful ones. Different from previous works, our method does not depend on pre-defined templates for context modeling and pays more attention to the useful context information attained by relation reasoning instead of taking equal treatment. This mitigates the ambiguity and thus can boost the performance of detection.

ARM3D is a plug-and-play module which can be conveniently applied to different 3D object detectors. It provides precise and useful relation context to help 3D detectors locate and classify objects more accurately and robustly. We apply ARM3D to two 3D object detectors and evaluate its improved performance on two challenging datasets. Extensive experiments demonstrate the effectiveness of ARM3D. Specifically, applying ARM3D to VoteNet~\cite{Qi_2019_ICCV} achieves  $\textbf{7.8}\%$ improvement on ScanNetV2~\cite{dai2017} and $\textbf{3.4}\%$ on SUN RGB-D dataset~\cite{song2015}. As for MLCVNet~\cite{Xie_2020_CVPR}, we achieve $\textbf{3.4}\%$ improvement on ScanNetV2. 

In summary, the major contributions of this paper are:
\begin{itemize}
\item a novel attention-based 3D relation module, using a simple but useful objectness module to perform object-aware relation reasoning between selected proposals, which can extract reliable and rich semantic and spatial relation contexts for detection.

\item an expressive attention module based on Transformer, intended to avoid the negative effects of confusing relation contexts and thereby enabling each object to take full advantage of the most useful context from others. Incorporated with the proposed objectness module and attention module, our method ARM3D can achieve more accurate and detection performance.

\item extensive experiments demonstrate the benefits of our attention-based relation module. Using our relation module in two state-of-the-art detectors shows substantial improvements on ScanNetV2 and SUN RGB-D benchmarks indicating that our design is effective and can be widely applicable.
\end{itemize}

\label{Relatedwork}
\section{Related Work}

\textbf{3D object detection in point clouds.} 3D object detection has been investigated for decades with numerous applications~\cite{Lin_2013_ICCV,shi2019hierarchy, Xie_2020_CVPR, Chen_2020_CVPR, Qi_2019_ICCV,H3DNet,Qi_2018_CVPR,Chen2017Multi,ku2018joint,shi2019pointrcnn,xu2012fit,xu2011photo, zhao2018triangle}. However, due to the orderless and sparse properties of point clouds, one of the main 3D representations, 3D object detection still remains challenging. Before the emergence of deep learning techniques on 3D point clouds~\cite{wang2017cnn,li2018pointcnn,atzmon2018point}, earlier attempts mainly turn to intermediate solutions such as using voxel grids~\cite{yan2018second,lang2019pointpillars,shi2020points}, multi-view images~\cite{pang20163d,Chen2017Multi} or trying to transform 2D object candidates to 3D from existing 2D object detection methods~\cite{Qi_2018_CVPR,lahoud20172d}, which limits the applicability in certain situations. 

Thanks to PointNet/PointNet++~\cite{Charles2017PointNet,qi2017pointnet++}, in recent years 3D object detection has started to take point clouds directly as input. Inspired by Faster RCNN~\cite{ren2015faster}, PointRCNN~\cite{shi2019pointrcnn} uses a two-stage 3D object detector for proposal generation and refinement. 
Li \emph{et al.} proposed GSPN~\cite{yi2019gspn}, a novel object proposal generation network by reconstructing shapes from noisy observations in a scene with an analysis-by-synthesis strategy. Motivated by Hough voting in 2D object detection, VoteNet~\cite{Qi_2019_ICCV} presents an end-to-end trainable 3D object detection framework and highlights the challenge and importance of directly predicting bounding box centers in point clouds because most surface points are far from the object centers. Extension works of VoteNet~\cite{Xie_2020_CVPR,Chen_2020_CVPR,yang20203dssd,Engelmann20CVPR,cheng2021brnet} make use of contextual information, graph neural networks with hierarchical structures, better reasonable sampling strategies and back-tracing representative cluster points for better proposal generation. In fact, explicit relationships between objects provide abundant information for scene understanding yet is ignored by these methods. The significance of relation contexts between objects for 3D box estimation is also emphasized by Huang \emph{et al.}~\cite{huang2018cooperative}.

\textbf{Relational reasoning in 3D.} With the emergence of the Relation Network~\cite{santoro2017simple}, there has been a great many methods that adapt the Relation Network~\cite{santoro2017simple} to various 2D image tasks~\cite{hu2018relation,mou2019relation,li2020spatial,chen2017spatial,cui2020learning,huang2020improving,Krishna2016Visual,liu2020beyond,cadene2019murel,hu2018relation,mou2019relation,sung2018learning,wang2019exploring}. The successful applications of these works illustrate the importance of relation reasoning in visual tasks. 

As a result of the successful applications of relational reasoning in 2D, various works began to explore its applications in 3D. For furniture layout in 3D,~\cite{huang2016structure} defines five types of relations for modeling furniture in indoor scenes using a graph structure, which, however, is time-consuming for relations like \emph{facing} and~\cite{song2017web3d} measures the similarity between various furniture layouts with case-based reasoning. Duan \emph{et al.}~\cite{duan2019structural} takes advantage of PointNet~\cite{Charles2017PointNet} to reason about the local structural dependencies with an additional relation network and attain improved performance in point cloud classification as well as part segmentation. Aimed at pose estimation,~\cite{kulkarni20193d} proposes a joint object and relation network to analyze the relative poses between each pair of objects. For 3D object detection, Xie \emph{et al.}~\cite{Xie_2020_CVPR} exploit self-attention to reason meaningful contextual information to generate better qualified proposals at three levels. GRNet~\cite{li2020grnet} propose a geometric relation network to leverage intra-object and inter-object features extracted by aggregation for 3D object detection.~\cite{lan20213drm} propose a relation module that explicitly defines the semantic and spatial relations between objects to get better relation contexts for 3D object detection. However, these works usually ignore the fact that part of the contextual information is misleading, and may degrade the performance in visual tasks when combined with correct information in complex environments.

\textbf{Attention in 3D vision.} Attention is an intelligent mechanism which can highlight what is important in a flexible manner. Recently, there have been numerous methods introducing attention to all kinds of 3D vision tasks.~\cite{wang2019graph,chen2019gapnet,wen2021airborne} intuitively leverage attention-based graph structures to capture the fine-grained features of 3D points for point cloud classification and segmentation.~\cite{wen2020point} proposes a skip-attention mechanisms to bridge local region features and point features of the decoder for better point cloud completion. There are also applications of attention mechanism in point cloud registration~\cite{wang2019deep,yew20183dfeat} and point cloud based retrieval~\cite{zhang2019pcan,sun2020dagc}. Moreover,~\cite{guo2021pct,zhao2020point} adapt Transformer, which attracts much attention in natural language processing, to 3D point cloud learning, and obtain high performance. Inspired by these methods, we utilize an expressive attention module mainly based on Transformer to model the importance of relation contexts of different object pairs for more accurate and robust 3D object detection.

\label{methods}
\section{Method}

\label{Overview}
\subsection{Overview}

Contextual relationships have been shown to be useful. However, there are still two main challenges when applying relational reasoning to 3D object detection. Firstly, most existing methods resort to object proposals first and rely on these proposals to perform bounding box classification and regression. Objectness of these raw proposals is usually represented as proposal quality, which actually makes a difference to relational reasoning. Proposals with low objectness, however, usually account for the majority, resulting in misleading context to some extent. Secondly,  even for high quality proposals, simply extracting the relation contexts between these proposals is not robust enough. Previous methods give relation contexts equal importance. This inevitably includes contradictory information with regard to a single object and may laed to ambiguity in 3D object detection.  

To overcome these two challenges and utilize relation contexts better, we have designed an attention-based relation module, \emph{ARM3D} for short, to distribute unequal attention towards relation contexts with different qualified object proposals. See Figure \ref{fig:framework}: with point cloud as input, different backbones can be used to generate numerous object proposals. By taking features of these proposals as input, ARM3D first selects proposals with high objectness scores through MLP which in itself enhances reliability, and then each proposal is matched with other proposals in the same scene at random. Moreover, ARM3D uses an attention module to model the importance of different relation contexts for each selected proposal. For proposal $p_i$, darker blue indicates greater importance. Both semantic and spatial relational reasoning is performed to extract the contextual relation features for more robust and accurate detection. 

In summary, we propose object-aware relational reasoning for the first challenge (see Section~\ref{relation reasoning}) and an attention module based on Transformer structures for the second challenge (see Section~\ref{Strategies for backbones}). Designs for loss function for ARM3D and its application to current 3D detectors are considered in Section~\ref{Loss Function}. Extensive experiments show that our design can not only achieve more accurate and robust detection performance but also mitigate the ambiguity in 3D object detection.

\begin{figure*}[htp]

  \centering
    \includegraphics[width=1.0\linewidth]{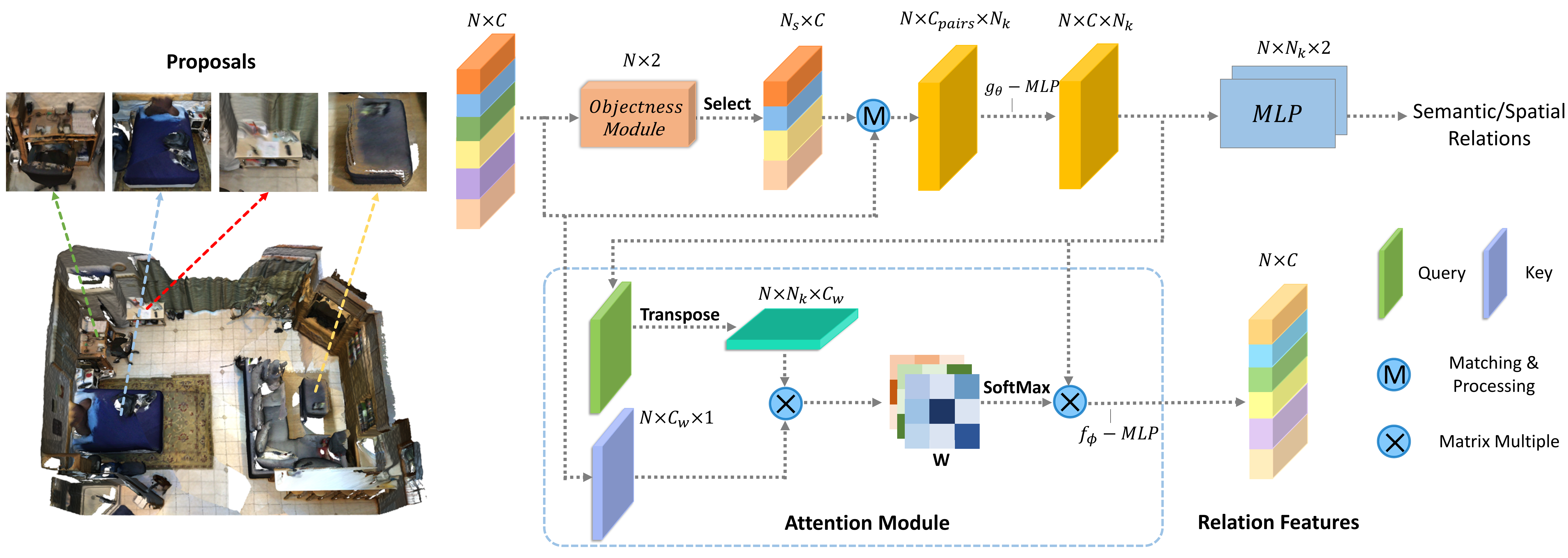}
  \caption{Network architecture of ARM3D. With $N$ proposals as input, the objectness module, mainly composed of MLPs, firstly outputs binary labels to select $N_s$ proposals with high objectness. $C$ indicates the feature channels. Each proposal is matched with a certain number of selected proposals at random, and further operations, including matrix subtraction and concatenation, are performed on these object pairs to obtain their differences. Pair-wise features corresponding to the same proposal go through the MLP labelled $g_{\theta}$. The extracted pairs of features are then transposed and fed into other MLPs to reason about semantic or spatial relations: see Section~\ref{relation reasoning}. Moreover, the original $N\times C$ proposals and pair-wise proposals go through two MLPs named \emph{Key} and \emph{Query} MLPs which output the matrices that are multiplied to compute the attention matrix. SoftMax activation follows, which is then multiplied by pair-wise features. Processed by the $f_\phi$ MLP, the relation module outputs relation features for each proposal.}
  \label{fig:fig_relation}       
  \end{figure*}

\subsection{Object-aware relational reasoning}
\label{relation reasoning}

Relation reasoning has been proven to be beneficial to 3D scene understanding~\cite{huang2018cooperative,huang2016structure}. In fact, objects in the same scene are typically related to each other. For instance, only half of a chair beside a table may be visible in a point cloud due to noise, but it is still likely to be recognized as a chair using human intuition. The reason why people can successfully understand this situation is that we know that chairs are often found beside tables in indoor scenes. This means that chairs are usually by the side of a table in indoor scenes. However, for neural networks, it is hard to model the correlation directly to provide prior information as used by humans. Thus, we need relational reasoning to model the high-level correlations between different objects involved in 3D object detection. We use two typical and intuitive relations including semantic and spatial relations to help the networks to learn the correlations. Furthermore, relational reasoning should be carried out for proposals with high objectness to avoid misleading contexts. As shown in Figure \ref{fig:fig_relation}, the upper part indicates the process of object-aware relation reasoning. Its components are as follows.

\textbf{Objectness module}. The reason why we need an objectness module to filter proposals is that poor quality proposals usually produce misleading contexts during relational reasoning. With $N\times C$ proposals in a given scene as input, the objectness module outputs $N\times 2$ binary labels demonstrating whether the proposals have high enough objectness to be qualified for relational reasoning. $C \in \mathbb{R}^{d}$ denotes the number of feature channels for each proposal generated by the backbones. Specifically, if the Euclidean distance $d_i$ between the center of a proposal $c_i$ and the center of its nearest ground-truth object is within a certain threshold $\xi$, the objectness label of this proposal is $1$, and $0$ otherwise:
\begin{equation}
  \label{equ:gt_obj}
  d_{i}= \min({D}(c_{i},c_{g})), g\in\lbrace1,\ldots,N_{gt}\rbrace
  \end{equation}  
where $D$ denotes Euclidean distance; $c_g$ is the center of a ground truth object, and $N_{gt}$ is the number of ground-truth objects in a scene.

The structure of objectness module $H_\varphi$ is compromises of three MLPs including $h_1, h_2, h_3$, with $C/2, C/4, 2$ output feature channels respectively, and each convolution layer is followed by batch normalization and ReLU activation. The output of the objectness module is a binary label which indicates whether the proposal is a single object or not. It can be formulated as Eq. (\ref{equ:objectness})

\begin{equation}
  \label{equ:objectness}
  l_{{obj}}  = \mathrm{argmax}(H_{\varphi}(p_{i})), i\in\lbrace1,\ldots,N\rbrace
  \end{equation}
where $l_{obj}$ denotes the objectness prediction of proposal $p_i$, and $H_{\varphi}(p_i)$ indicates the binary logits of the last layer.

\textbf{Matching and processing}. Since the objectness module provides an objectness prediction for each proposal, it is simple to select $N_s\times C$ high quality proposals. We argue that pair-wise relation context benefits the detection of each proposal to the full extent if the context is extracted from proposals with high objectness. Each proposal is matched with $N_k$ proposals among the $N_s$ ones selected by the objectness module in the same scene at random. The strategy of random matching is intended to increase the diversity of object-wise relation contexts, while the attention module in Section \ref{Strategies for backbones} is able to choose the more useful ones. Instead of using sampling strategies like farthest point sampling (FPS) or $k$-nearest neighbors ($k$-NN) to select proposals for matching, we argue that the selection results are relatively unchanged for these sampling methods. Modeling the accurate correlation between proposals is a great challenge and we use random sampling to increase the diversity for better understanding since we have an attention module to keep the relation contexts stable and useful. In our experiments, other sampling strategies work less well than random sampling. To process a pair of proposals ($p_i, p_j$), from the features of $p_i$ we subtract those of $p_j$ to obtaining the difference, which is concatenated with $p_i$, and formulated as features of proposal pairs $N\times C_{pairs}\times N_{k}$.

To decide on the informativeness of context provided by proposal pairs, we leverage MLPs called $g_{\theta}$ to exploit the semantic or spatial relations within them. These pair-wise features are sent to the classification MLPs named $R_{\theta}$ to predict their semantic or spatial relation labels. For proposal $p_i$, the relation label $l_r$ of itself and its matched proposal $p_{j}$ can be formulated as follows:

\begin{equation}
  \label{equ:rn}
  l_{r}  = R_{\theta}(g_{\theta}(C_{\psi}(p_{i},\Delta(p_{i},p_{j}))))  ,p_{j}\in{P_k}
  \end{equation}
where $C_{\psi}$ denotes the concatenation of features and $\Delta$ indicates subtraction. $P_k$ denotes the randomly matched proposals for $p_{i}$.

\begin{figure*}[htb]
  \centering
  \includegraphics[width=1.0\textwidth]{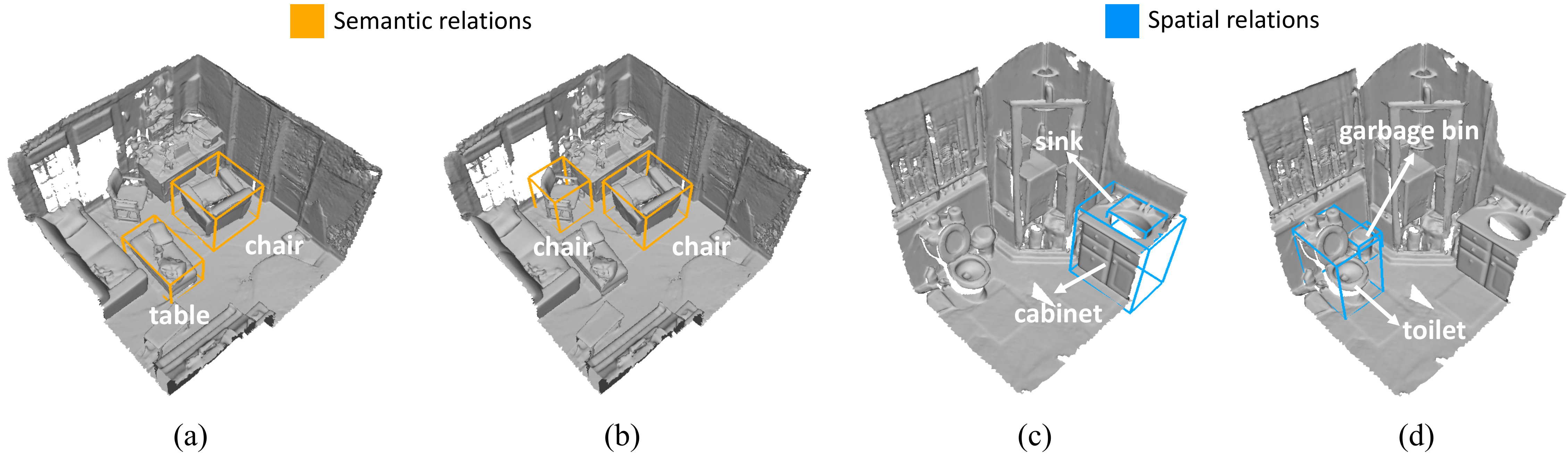}
  \caption{Semantic and spatial relations. The orange bounding boxes indicate semantic relations, and the blue bounding boxes show spatial relations. (a) Semantic relations in different categories between a chair and the table. (b) Semantic relations in the same category between these two chairs. (c) Vertical spatial relations between the sink and the cabinet. (d) Horizontal spatial relations between the toilet and the garbage bin beside it. Best viewed on screen.}
  \label{fig:vis-relation_render}
\end{figure*}

\textbf{Semantic and spatial relations}. Motivated by Relation Networks proposed in~\cite{hu2018relation}, \cite{lan20213drm} adapts it to 3D object detection and explicitly performs relational reasoning on individual objects instead of on the entire scene. The main differences between our method and \cite{lan20213drm} are that we simplify the semantic relations to exclude relations between the same instance, and we use an attention module combined with an objectness module to make full use of the relation contexts to avoid redundant contexts and provide better performance. In this paper, the original relations presented in \cite{lan20213drm} including \emph{group, same as, support, hang on} is simplified. Since only proposals with high objectness are selected for extracting relation contexts, relations like \emph{same as} which indicates that two proposals belong to the same instance are in a minority and unsuitable in this case. Therefore, we believe that semantic relations and spatial relations are the most typical and beneficial pair-wise object relation types for indoor 3D object detection, which are exactly sufficient for gathering nontrivial relation contexts. As shown in Figure~\ref{fig:vis-relation_render}, two types of relations indicate whether two objects are in the same category or not, and whether one is linked to the other horizontally or vertically.

With regard to semantic relations, there are usually various types of objects in indoor scenes. If two objects are in the same category, the semantic relation label is 1, and 0 otherwise. For each object, distinguishing semantic categories from other objects implies rich semantic information. The goal of semantic relations is to capture the semantic class-specific properties between objects. Objects in the same category usually have similar structures and parts, which helps the objects to better recognize themselves with semantic context. In contrast, for objects of different categories, an object can learn differences from their structures and appearance through the semantic relations. Although the principle of semantic relations is simple, it is useful and informative for 3D object detection.

As for spatial relations, we combine the relations of~\emph{support, hang on} proposed in~\cite{lan20213drm} together as \emph{spatial} relations. In this paper, spatial relations indicate that two objects are adjacent to each other horizontally or vertically, which can indicate that one is supporting or linked to the other one. In reality, objects are more or less spatially related, especially for those with 3D representations. For example, a chair is under a table, or a bookshelf is beside a wall. Such cases are typical spatial relations for indoor object pairs, which are the intuitional and give intuititive and meaningful contexts for object detection. We define that a spatial relation exists only if two proposals satisfy two conditions. First, the relative height $H_{r}$ or horizontal distance $L_{r}$ of two proposals should be lower than a threshold $\tau_{d}$. This means that the nearest distance between points of two proposals should be small enough, either horizontally or vertically. Second, the overlapping ratio of bounding boxes for two proposals should be larger than a threshold $\tau_{r}$ either on the $x$-$y$ plane, the $y$-$z$ plane and the $z$-$x$ plane with respect to the first condition. Take the $x$-$y$ plane as an example. The overlap ratio $r_{i,j}$ can be calculated as:
\begin{equation}
  \label{equ:spatial}
  r_{i,j}  = \max(\frac{\varOmega_{xy}(p_i,p_j)}{\varphi_{xy}(p_i)}, \frac{\varOmega_{xy}(p_i,p_j)}{\varphi_{xy}(p_j)}) 
  \end{equation}
where $\varOmega_{xy}(\cdot, \cdot)$ denotes the area of intersection in projection for two proposals on the horizontal plane; $\varphi_{xy}(\cdot)$ is the projected area of a proposal on the horizontal plane.

If the overlap ratio $r_{i,j}$ is lower than $\tau_{r}$, the pair of proposals $(p_i, p_j)$ is supposed to have spatial relations, similarly for the $y$-$z$ plane and the $z$-$x$ plane. Such compact spatial relations are helpful for 3D object detection as well as scene understanding.

\begin{algorithm}[t]
  \begin{algorithmic}
      \STATE \textbf{Input:} Proposal $p_i$, $N_k$ pairs of proposals $(p_{i}, p_{j})$ and MLPs $f_{\phi}, Key, Query$.
      \STATE \textbf{Output:} Weighted relation features $R_i$.
      \STATE \textbf{Initialize:} $R_{i}=0$, $W=0$.
      \FORALL{$j\in \lbrace1,\ldots,N_{k}\rbrace$}
      \STATE matrix $K = $ Key($p_i$), matrix $Q =$  Query($(p_{i}, p_{j})$);
      \STATE $Q = \tanh(Q), K = \tanh(K)$;
      \STATE $W_{i,j} = Q^T\times K$ and $W\leftarrow W_{i,j}$.
      \ENDFOR
      \STATE Normalize $W$ matrix by SoftMax; $j=0$.
      \WHILE{$j\leq N_k$}
      \STATE $R_{tmp} = f_{\phi}(W_{i,j}\times (p_i,p_j))$;
      \STATE $R_i = R_i + R_{tmp}$; $j = j+1$
      \ENDWHILE
      \STATE Return the weighted relation features $R_i$.
  \end{algorithmic}
  \caption{Pseudo-code for attention-based relation features formulation.
  }\label{algorithm1}
\end{algorithm}

\subsection{Attention module}
\label{Strategies for backbones} 
Although relation contexts are generally beneficial for detection, not all contexts from other objects are essential and helpful for a single object. It is common and inevitable that some pair-wise relation contexts are misleading and even useless for specific objects (see Figure~\ref{fig:teaser}). The Attention mechanism, which has become a focus in 3D vision recently, is appropriate for solving this problem.

In order to make our relation module more expressive and robust, we adapt the attention module based on Transformer in~\cite{guo2021pct} for analyzing the importance of different pair-wise relation contexts for every single object. Unlike Transformer in~\cite{guo2021pct} which leverages self-attention to extract features of point clouds, our attention module is designed to weigh different pair-wise relation contexts. As shown in Figure~\ref{fig:vis-relation_render}, to be specific, the $N\times C$ original proposals first go through MLPs named \emph{Key} and the feature channel is downsampled to $C_w$. A similar operation is performed on $N\times C\times N_k$ pairs of proposals that are matched with the original $N$ proposals. After the MLPs called \emph{Query}, the pairs of proposals are transposed into $N\times N_k\times C_w$, and multiplied by the \emph{key} proposals to obtain the $N\times N_k$ attention matrix. Note that we use $\tanh$ activation to normalize the outputs before multiplication. Each row of the attention matrix corresponds to a proposal in the scene. Values in each row give the importance of relation contexts from different pairs of proposals, respectively. After softMax normalization, the attention matrix is used to assign different weights to the $N\times C\times N_k$ pairs of proposals; the sum of these values is used to compute the weighted average relation contexts. Last, the weighted relation context for each original proposal is fed into MLPs called $f_{\phi}$ to output the final relation features. The process of obtaining the relation features $R_i$ of proposal $p_i$ can be formulated as follows:

\begin{equation}
  \label{equ:rn_feature}
  W = \Theta(\Gamma(Q^T)\times \Gamma(K))
  \end{equation} 

\begin{equation}
  \label{equ:rn_feature}
  \textbf{R}_{i}  = f_\phi(\sum_{\forall_{j}} W_{i,j}\times (p_{i},p_{j})), j\in\lbrace1,\ldots,N_{k}\rbrace
  \end{equation} 
where $Q$ is the \emph{Query} output matrix and $K$ is the \emph{Key} output matrix; $\Gamma$ denotes the $\tanh$ activation function; $\Theta$ is softMax normalization, and $W$ indicates the attention matrix of different pairs of proposals $(p_{i},p_{j})$. Further details are provided in Algorithm~\ref{algorithm1}.

\subsection{Application and loss function}
\label{Loss Function}
To examine the effectiveness of our method, we have applied our attention-based relation module ARM3D to two state-of-the-art 3D object detectors: VoteNet~\cite{Qi_2019_ICCV} and MLCVNet~\cite{Xie_2020_CVPR}. Taking the grouped clusters as proposals, ARM3D predicts the pair-wise semantic or spatial relations between those with high objectness and outputs the beneficial relation features to boost the performance of 3D object detection.

The loss of ARM3D is simply made up of the objectness loss as well as the relation prediction loss, corresponding to Section~\ref{relation reasoning} and Section~\ref{Strategies for backbones} respectively. The objectness loss is formulated as $\mathcal{L}_{obj}$, which is used to supervise the module to predict the accurate objectness of each proposal. The relation prediction loss is formulated as $\mathcal{L}_{rn}$, which refers to the prediction loss of semantic or spatial relations between proposal pairs, using the binary cross entropy. For the better selection of objectness, we set different weights: $w_0$ for those proposals whose ground-truth objectness labels are false, and $w_1$ for the true ones. Similar strategies are adopted for $\mathcal{L}_{rn}$ too.
$\mathcal{L}_{rn}$ represents the loss for a single type of relation (semantic or spatial relations). The final relation loss $\mathcal{L}_{r}$ is the sum of these two losses. $\mathcal{L}_{rn}$ is formulated as follows:
\begin{equation}
  \begin{split}
  \mathcal{L}_{rn} = - \frac { 1 } { N_p } \sum _ { i = 1 } ^ { N_p } w_1 \cdot y _ { i } \cdot \operatorname { log } ( p ( y _ { i } ) ) \\
  + w_0 \cdot ( 1 - y _ { i } ) \cdot \operatorname { log } ( 1 - p ( y _ { i } ) )
  \end{split}
  \label{equ:lrn}
  \end{equation}
where $N_p$ is the number of proposal pairs with $N_p = N \times N_k$ in this paper. $y_i$ indicates the positive ground-truth semantic or spatial relation label of the proposal pair and $p{(y_i)}$ is the predicted possibility of the relation of this pair to be positive. $w_0$ and $w_1$ are  the weights as above.

Previous methods only calculate the objectness loss of proposals that are either within a small distance or beyond a large distance. Since the accuracy of objectness prediction makes a difference to our method, we calculate the objectness loss for all proposals and assign more weight to positive instances while training.

Following~\cite{Qi_2019_ICCV,Xie_2020_CVPR}, when using our ARM3D, the network is trained in an end-to-end manner by using a voting loss $\mathcal{L}_{{vote}}$, a 3D bounding box regression loss $\mathcal{L}_{{box}}$, and a semantic classification loss $\mathcal{L}_{{cls}}$, in addition to the objectness loss $\mathcal{L}_{{obj}}$ and relation prediction loss. The overall 3D object detection loss is formulated as: 
\begin{equation}
  \mathrm{loss} = \lambda_1\mathcal{L}_{vote} + \lambda_2\mathcal{L}_{{obj}} + \lambda_3\mathcal{L}_{{box}} + \lambda_4\mathcal{L}_{{cls}} + \lambda_5\mathcal{L}_{{r}} 
  \label{equ:loss}
  \end{equation}
where in our experiments, we set $\lambda_1=1.0, \lambda_2=0.5, \lambda_3=1.0, \lambda_4=0.1, \lambda_5=0.1$.

\label{implementation}
\section{Implementation Details}

In this section, we first describe the implementation details about the network architecture and the corresponding parameters for ARM3D. Then we explain how to apply our ARM3D to two 3D object detectors, VoteNet~\cite{Qi_2019_ICCV} and MLCVNet~\cite{Xie_2020_CVPR}, as well as the overall training strategies.

\textbf{Details in ARM3D.} Proposals are object candidates for 3D object detection. Our ARM3D selects proposals with an objectness module, and relation contexts can be extracted from these relatively reliable ones. For the ground-truth objectness labels, we set the distance threshold $\xi = {0.3}$. Objectness of proposals within $\xi$ with respect to their ground-truth objects is set to 1. Unlike previous methods that only compute the objectness loss of proposals within the \emph{near} distance threshold or the \emph{far} threshold, we focus on the objectness of all proposals. For $\mathcal{L}_{{obj}}$, we use the binary cross-entropy loss with different weights of $w_0 = 0.2$ and $w_1 = 0.8$ for the negative or positive cases respectively, since the backbone network initially produces few sufficiently good proposals. The same strategies and designs are applied to $\mathcal{L}_{{rn}}$ since there are relatively fewer positive samples.

As for the strategies of matching different proposals to obtain pair-wise features, we randomly choose $N_k = {8}$ proposals from the ones selected by the objectness module for the ScanNetV2 dataset and SUN RGB-D datasets. This strategy provides a good trade-off between speed and results. Moreover, random matching can diversify the relation contexts. For matched proposal pairs, we set the distance threshold $\tau_{d} = 0.1$ and the ratio threshold $\tau_{r} = 0.5$ to compute the ground-truth spatial relation labels.

For the attention mechanism used in ARM3D, the \emph{Query} and \emph{Key} MLPs are both composed of one convolutional layer to downsample the input feature from $C$ to $C_{w}=C/4$ followed by a $\tanh$ activation function. Different from other methods, these two MLPs do not share weights. The function $f_\phi$ is a fully connected layer with $C$ channels as output.

The computational requirements of our method are indicated in Table \ref{tab:complexity}. We compare VoteNet~\cite{Qi_2019_ICCV} to VoteNet using our ARM3D and the model size of our method is 14.2MB. The inference time using our method is 0.14s and 0.09s on ScanNetV2 and SUN RGB-D datasets, which is comparable to that for VoteNet alone. This demonstrates the efficiency of our method as a lightweight but useful plug-and-play module. The complexity of the calculation of semantic and spatial relations is relatively low-cost since we use matrix multiplication instead of loops in experiments.

\begin{table}[!t]
    \centering
    \caption{Model size and processing time (per frame or scan) for VoteNet, and VoteNet with our method ARM3D.}
    \scalebox{1.00}{
    \setlength{\tabcolsep}{0.5mm}
    \renewcommand\arraystretch{1.1}{
    \begin{tabular}{l|c|c|c}
    \toprule
    Method  &Model size & ScanNetV2 & SUN RGB-D     \\\hline
    VoteNet              & 11.2MB      & 0.12s  & 0.08s\\ \hline
    VoteNet+ARM3D   & 14.2MB     & 0.14s  & 0.09s\\ 
    \bottomrule
    \end{tabular}}}
    \label{tab:complexity}
\end{table}

\textbf{Details in Training.} We apply our ARM3D relation module to VoteNet~\cite{Qi_2019_ICCV} and MLCVNet~\cite{Xie_2020_CVPR} to examine whether our method is effective and widely applicable. The number of feature channels $C$ of proposals generated by these two methods is 128. Generally, we keep the same training strategies, including the base learning rates, decay steps, max training epochs and so on as in the original papers~\cite{Qi_2019_ICCV,Xie_2020_CVPR}. The only difference is that, when applied to VoteNet on ScanNetV2, the maximal training epoch is 180, and the batch size is kept as 4 for the first 80 epochs while the batch size is changed to 8 for the remaining epochs. For MLCVNet, we keep the batch size as $b = 8$ from beginning to end. We implement our approach using PyTorch~\cite{paszke2019pytorch} on a single NVIDIA TITAN V. During training, we find that the mAP results fluctuate slightly, so mAP results given here are mean results over three runs.

\label{results}
\section{Experiments}

In this section, we evaluate the proposed attention-based relation module ARM3D applied to two 3D object detectors, VoteNet~\cite{Qi_2019_ICCV} and MLCVNet~\cite{Xie_2020_CVPR}. With point clouds of indoor scenes as input, the experiments are performed on two large 3D indoor scene datasets and evaluated on the corresponding detection benchmarks. (see Section~\ref{Experimental dataset}). The evaluation metric we use is demonstrated in Section~\ref{Evaluation metric}. In Section~\ref{Evaluation on two detectors}, we analyze the performance improvement after applying our attention-based relation module ARM3D to the above two 3D object detectors. An ablation study for different components of our method is performed mainly with VoteNet on ScanNetV2 dataset (see Section~\ref{Ablation study}). Note that VoteNet depends on Deep Hough Voting for object detection, while MLCVNet extends VoteNet with additional three-level useful contexts; it is challenging for the effectiveness of the relation contexts from our ARM3D. Experimental settings are the same when applying our ARM3D to these detectors. Both quantitative and qualitative results show the effectiveness and generalization ability of our ARM3D.

\subsection{Dataset and benchmarks}
\label{Experimental dataset}
We use two widely used datasets that provide 3D point clouds of indoor scenes to evaluate our methods: ScanNetV2~\cite{dai2017} and SUN RGB-D~\cite{song2015}.

ScanNetV2 is a large RGB-D 3D indoor scene dataset with densely annotated 3D reconstructed meshes. There are approximately 1.5K scanned indoor scenes where both the semantic segmentation and bounding boxes of objects are given. The scanned indoor scenes are relatively complete, which makes it suitable for our method to extract the relation context.

SUN RGB-D is a well-known public single-view RGB-D dataset for scene understanding, which contains about 10K RGB-D images. The images are captured by four different sensors, providing accurately annotated oriented bounding boxes in 37 categories. Since it does not provide reconstructed point clouds, we convert the depth images to point clouds using known camera parameters. Most scenes are captured in household environments. Occlusion is common in the SUN RGB-D dataset, and there are fewer ground-truth objects in each scene, making it quite challenging for 3D object detection as well as relational reasoning.

\subsection{Evaluation metric}
\label{Evaluation metric}
The evaluation metric we take is the average precision of the detected object bounding boxes against those of ground-truth objects. We use two $ IoU $ thresholds of 0.25 and 0.5, respectively, in our experiments. The mean average precision mAP is the macro-average of the average precision across all test categories.

\begin{figure*}[!t]
    \centering
    \includegraphics[width=0.98\textwidth]{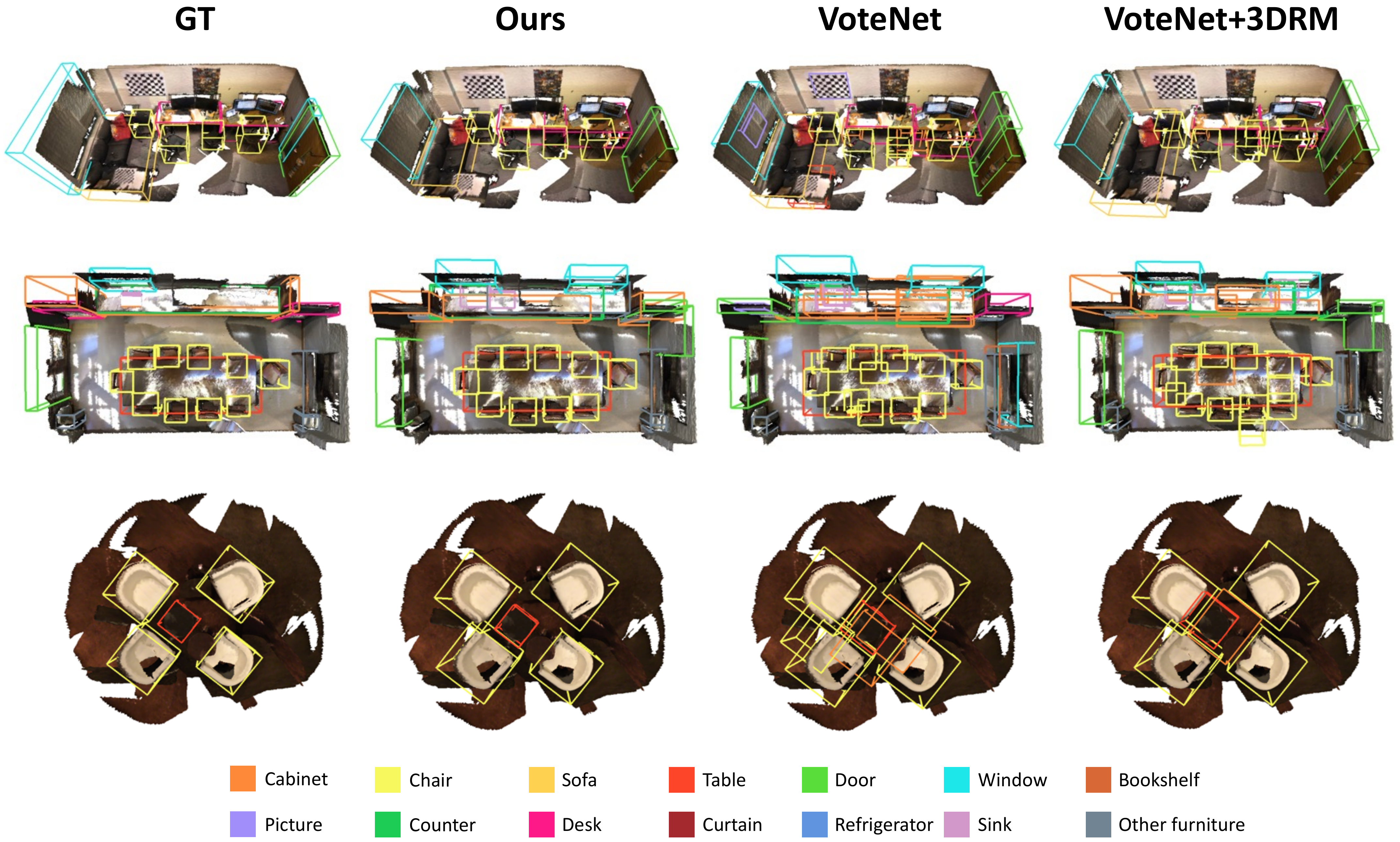}
    \caption{Qualitative comparison results of 3D object detection on the ScanNetV2 val set. Columns left to right: ground-truth, our method,  VoteNet, VoteNet+3DRM. The detailed comparison demonstrates that our method ARM3D enables more accurate and reasonable detection.}
    \label{fig:vis-scannet1}
\end{figure*}

\subsection{Evaluation on two detectors}
\label{Evaluation on two detectors}
\subsubsection{Overview}
We apply our ARM3D to two 3D object detectors, which are VoteNet~\cite{Qi_2019_ICCV} and MLCVNet~\cite{Xie_2020_CVPR} respectively. These two methods are regarded as our baselines to examine the effectiveness and improvements of our method ARM3D. We also compare the effects of applying 3DRM~\cite{lan20213drm} to these two detectors. We first analyze the improvement of VoteNet equipped with our ARM3D, which is denoted \textbf{VoteNet+ARM3D}. Then we analyze the increased performance after applying our ARM3D to MLCVNet, which is denoted as \textbf{MLCVNet+ARM3D}. A brief introduction to these two baselines and a pair-wise relation module for 3D object detection are given below.

\begin{table}[!t]
    \centering
    \caption{Comparison of our approach against VoteNet and MLCVNet on 3D object detection on ScanNetV2 and SUN RGB-D val sets. VoteNet+3DRM and MLCVNet+3DRM* use 3DRM~\cite{lan20213drm}. VoteNet+ARM3D and MLCVNet+ARM3D indicate VoteNet and MLCVNet equipped with our ARM3D.}
    \scalebox{0.85}{
    \setlength{\tabcolsep}{0.7mm}
    \renewcommand\arraystretch{1.1}{
    \begin{tabular}{l|cc|cc}
    \toprule
    \textbf{}   &\multicolumn{2}{c|}{ScanNetV2}      &\multicolumn{2}{c}{SUN RGB-D} \\
    \textbf{}		& mAP@0.25 & mAP@0.5  & mAP@0.25 & mAP@0.5\\ \hline 
    VoteNet      & 58.6  & 33.5  & 57.7  & 33.7 \\  
    VoteNet+3DRM & 59.7  & 37.3 & 59.1 & 35.1\\  
    VoteNet+ARM3D & \textbf{62.6}  & \textbf{41.3} & \textbf{59.3} & \textbf{37.1}\\ 
    \hline 
    MLCVNet & 64.5  & 41.4 & 59.8 & - \\  
    MLCVNet+3DRM$^*$ & 63.6  & 40.2 &  58.4 & 34.3  \\  
    MLCVNet+ARM3D & \textbf{64.8}  & \textbf{44.8} & \textbf{60.1}  & \textbf{35.8}   \\  
    \bottomrule
    \end{tabular}
    \label{tab:all_table}}}
\end{table}

\begin{itemize}
    \item \textbf{VoteNet~\cite{Qi_2019_ICCV}:} An end-to-end trainable 3D object detection framework that takes advantage of deep Hough voting and aggregation to generate proposals for scenes. The aggregated clusters are used to perform classification and bounding box regression.
    
    \item \textbf{MLCVNet~\cite{Xie_2020_CVPR}:} A method that utilizes three levels of implicit contexts to enhance the performance of VoteNet, including patch-wise, object-wise and global contexts.
    \item \textbf{3DRM~\cite{lan20213drm}:} A pair-wise plug-and-play relation module for 3D object detection, which takes advantage of four types of relations to improve the performance of 3D object detectors.
\end{itemize}

\begin{figure*}[!t]
  \centering
  \includegraphics[width=0.98\textwidth]{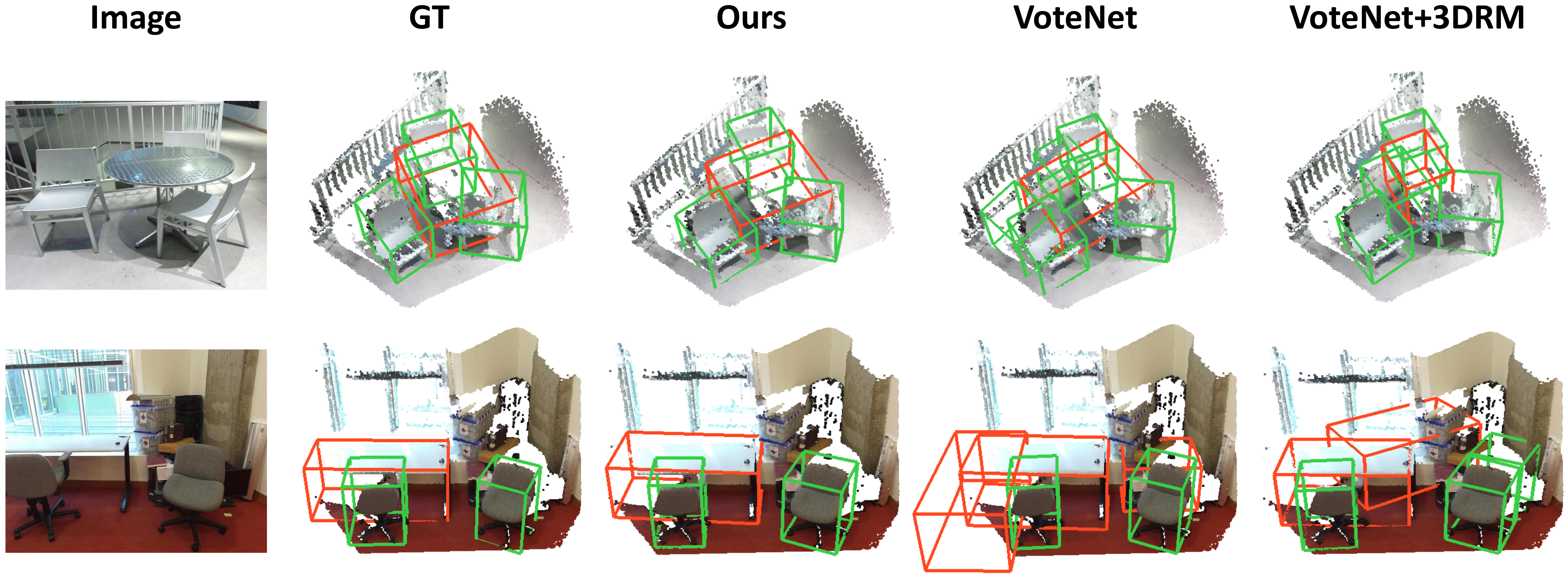}
  \caption{Qualitative comparison results of 3D object detection on SUN RGB-D val set. Columns left to right: RGB image of the scene, ground-truth, our method, VoteNet, VoteNet+3DRM. Our method VoteNet+ARM3D provides better results. Color is for depiction, not used for detection.} 
  \label{fig:vis-sunrgbd1}
\end{figure*}

\subsubsection{Comparison to baselines.} We evaluate our method against VoteNet, MLCVNet, and methods of applying 3DRM~\cite{lan20213drm} to these detectors. Table \ref{tab:all_table} reports the average precision on the ScanNetV2 and SUN RGB-D datasets with mAP@0.5 and mAP@0.25 respectively. Our method \textbf{VoteNet+ARM3D} , and \textbf{MLCVNet+ARM3D} achieves the best performance on ScanNetV2 val set and SUN RGB-D val set both.

\begin{table*}[!t]
    \centering
    \caption{Comparison to VoteNet and VoteNet+3DRM with mAP@0.5 on ScanNetV2 val set for our method with different relations. We denote VoteNet+ARM3D as VoteNet equipped with our ARM3D.}
    \scalebox{0.83}{
    \setlength{\tabcolsep}{0.82mm}
    \renewcommand\arraystretch{1.3}{
    \begin{tabular}{l|cccccccccccccccccc|c}
    \toprule
    \textbf{}	& wind	& bed  & cntr & sofa & tabl & showr & ofurn & sink & pic & chair & desk & curt & fridge & door & toil & bkshf & bath & cab & mAP\\  \hline
    VoteNet & 6.4 & 76.1 & 9.5 & 68.8 & 42.4 & 10.0 & 11.7 & 16.8 & 1.3 & 67.2 & 37.5 & 11.6 & 27.8 & 15.3 & 86.5 & 28.0 & 78.9 & 8.1 & 33.5 \\
    \hline
    VoteNet+3DRM  & 12.3  & 80.6  & 14.6  & 71.8  & 41.3  & 10.4  & 13.4  & 29.5  & 0.1  & 67.7  & 34.7  & 17.0  & 37.8  & 15.7  & 90.0  & \textbf{44.2}  & \textbf{83.0}  & 8.0  & 37.3 \\ 
    \hline
    VoteNet+ARM3D (semantic) & 10.3 & \textbf{82.4} & \textbf{32.0} & \textbf{76.2} & 51.9 & 14.4 & \textbf{20.9} & \textbf{32.3} & 0.2 & 75.0 & \textbf{48.4} & 14.7 & \textbf{40.2} & 20.0 & 85.9 & 40.9 & 77.6 & 12.9 & 40.9 \\

    VoteNet+ARM3D (spatial) & \textbf{12.9} & 80.7 & 24.1 & 73.5 & \textbf{55.1} & 11.7 & 20.6 & 28.1 & 2.1 & \textbf{76.5} & 43.7 & 17.7 & 36.2 & 20.3 & 85.3 & 36.6 & 77.5 & 14 & 39.8 \\ 
    VoteNet+ARM3D (all) & 9.1 & 78.2 & 28.2 & 71.2 & 54.0 & \textbf{24.4} & 18.7 & 25.9 & \textbf{2.8} & 75.6 & 44.1 & \textbf{23.9} & 37.6 & \textbf{21.9} & \textbf{92.0} & 43.1 & 79.1 & \textbf{13.4} & \textbf{41.3}\\

    \bottomrule
    \end{tabular}}}
    \label{tab:votenet-relations-05}
\end{table*}

From the comparison in Table \ref{tab:all_table}, our method significantly outperforms VoteNet by not only \textbf{4\%} and \textbf{7.8\%} on ScanNetV2 but also \textbf{1.6\%} and \textbf{3.4\%} on SUN RGB-D for mAP@0.25 and mAP@0.5 respectively. Note that MLCVNet+3DRM$^*$ means that we retrain 3DRM~\cite{lan20213drm} on MLCVNet since 3DRM does not have this application. Compared to applying 3DRM to VoteNet, our method outperforms it by $2.9\%$ and $4\%$ on ScanNetV2 as well as by $0.2\%$ and $2\%$ on SUN RGB-D for mAP@0.25 and mAP@0.5 respectively. This shows that our attention-based relation module can extract more robust and accurate relation contexts to benefit the 3D object detectors for better classification and regression. Note that the increased performance on SUN RGBD val dataset is slightly lower than on the ScanNetV2 validation dataset, since SUN RGBD is a single-view RGB-D dataset. Most scenes in the SUN RGB-D dataset are in household environments, and have fewer objects. Occlusion is more common in SUN RGB-D dataset than in ScanNetV2, making it quite challenging for detection as well as extracting relation contexts for our method. However, Table \ref{tab:all_table} illustrates that our method ARM3D can reliably reason about the relational context even in challenging scenes and environments.

\begin{table}[!t]
    \centering
    \caption{Comparison of our approach with different components against the baseline of VoteNet+3DRM on ScanNetV2 val set. We denote OBM as the objectness module and ATM as the attention module, respectively. VoteNet+ARM3D indicates applying our method ARM3D to VoteNet. Note that we only utilize the semantic relations in this experiment.}
    \scalebox{1.03}{
    \setlength{\tabcolsep}{0.65mm}
    \renewcommand\arraystretch{1.1}{
    \begin{tabular}{l|cc|cc}
    \toprule
    \multirow{2}{*}{Method} & \multirow{2}{*}{OBM} & \multirow{2}{*}{ATM} &\multicolumn{2}{c}{ScanNetV2}      \\
                  &        &          & mAP@0.25  & mAP@0.5\\ \hline 
    Baseline      &          &        & 59.7      & 37.3 \\  
    VoteNet+ARM3D & $\surd$  &        & 60.9      & 38.7\\  
    VoteNet+ARM3D &  & $\surd$ & 61.5      & 37.8\\  
    VoteNet+ARM3D & $\surd$ & $\surd$ & \textbf{62.9}      & \textbf{40.9}\\  
    \bottomrule
    \end{tabular}
    \label{tab:ablation_method}}}
\end{table}

While MLCVNet uses three levels of contexts to boost its performance, our method ARM3D can still improve its performance on 3D object detection via fine-grained relation contexts from ARM3D. Equipped with ARM3D, our method improves MLCVNet by \textbf{0.3\%} and \textbf{3.4\%} on ScanNetV2 for mAP@0.25 and mAP@0.5 respectively. Our method also outperforms MLCVNet by \textbf{0.3\%} on SUN RGB-D in terms of mAP@0.25. In contrast, applying 3DRM~\cite{lan20213drm} to MLCVNet reduces the performance of MLCVNet due to the equal weights towards relation contexts from different proposal pairs, which may contain some misleading contexts. It is noteworthy that our method ARM3D still can improve the performance of MLCVNet which already fuse various contexts to help the detection while 3DRM cannot do this. This further explains the effectiveness and universal benefits of relation contexts extracted by our ARM3D.

\begin{figure}[!t]
    \centering
    \includegraphics[width=0.49\textwidth]{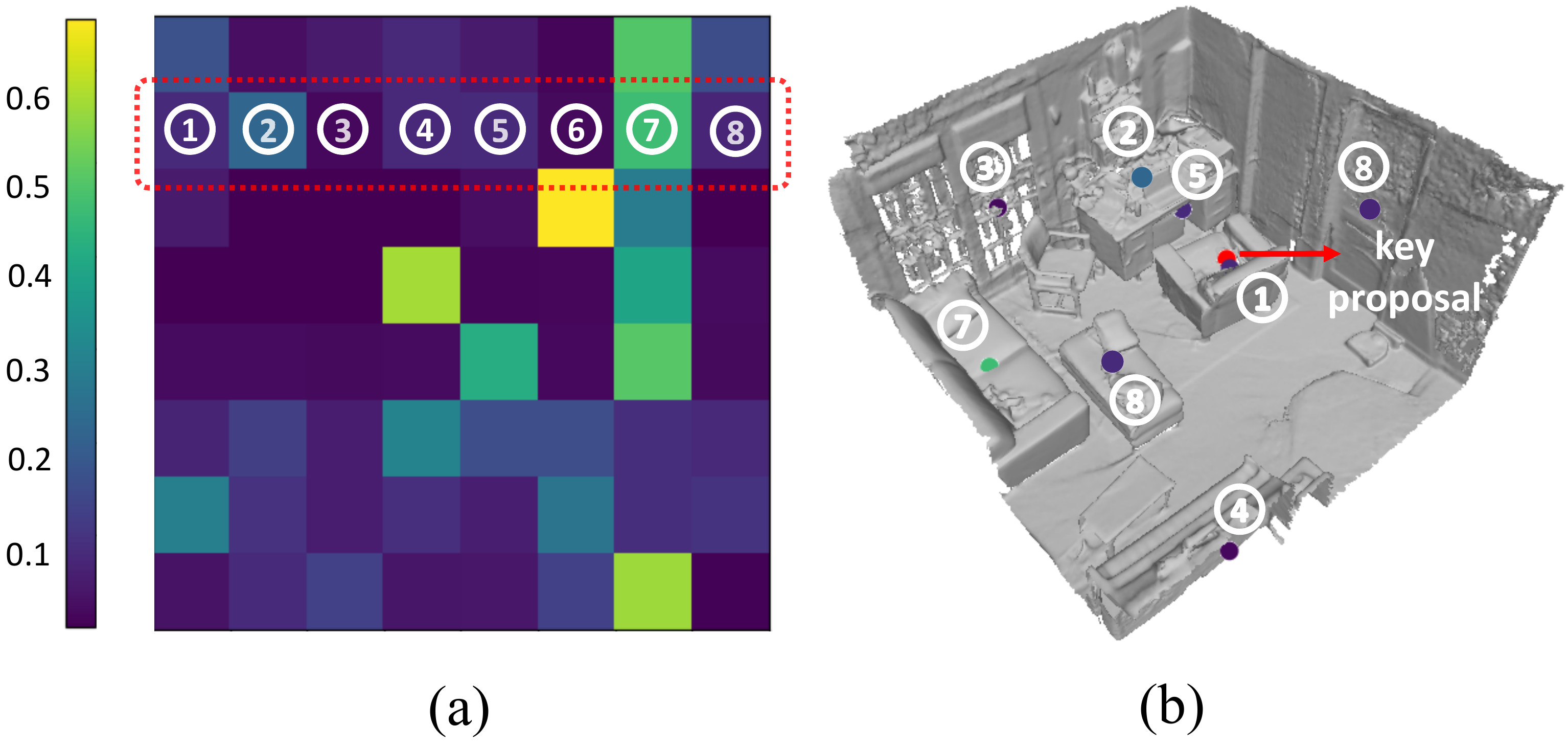}
    \caption{Attention in VoteNet+ARM3D. (a) $8\times 8$ attention matrix. Each row represents a proposal and corrsponding columns represent weights of other proposals towards it. (b) visualization of the second row in (a), which is numbered 1--8 as for the key proposal (the red dot). The other eight proposals are shown in dots with weighted colors.}
    \label{fig:vis-attention}
\end{figure}

\begin{table}[!t]
    \centering
    \caption{Comparison of our ARM3D with different numbers $N_k$ of proposals pairs for each one while relational reasoning on the ScanNetV2 val set. We denote VoteNet+ARM3D as our approach by applying our ARM3D on VoteNet.}
    \scalebox{1.05}{
    \setlength{\tabcolsep}{0.65mm}
    \renewcommand\arraystretch{1.12}{
    \begin{tabular}{l|cc}
    \toprule
    \multirow{2}{*}{Method}  &\multicolumn{2}{c}{ScanNetV2}      \\
                          & mAP@0.25  & mAP@0.5\\ \hline 
    VoteNet              & 58.6      & 33.5 \\ \hline
    VoteNet+ARM3D($N_k=2$)    & 61.5      & 38.8\\  
    VoteNet+ARM3D($N_k=4$)  & 61.3      & 39.8\\  
    VoteNet+ARM3D($N_k=6$)   & 62.0     & 39.7\\
    VoteNet+ARM3D($N_k=8$)   &\textbf{62.9}   &\textbf{40.9}\\  
    VoteNet+ARM3D($N_k=12$)   & 61.2      & 38.5\\  
    VoteNet+ARM3D($N_k=16$)   & 60.6     & 37.9\\  
    \bottomrule
    \end{tabular}
    \label{tab:ablation_pairs}}}
\end{table}

\subsubsection{Qualitative results and discussion} Qualitative results for different methods and ground truth for ScanNetV2 are shown in Figure~\ref{fig:vis-scannet1}. We visualize the results of ground-truth (the first column), our method (the second column), VoteNet (the third column) and VoteNet+3DRM (the last column). Thanks to ARM3D, our method obviously detects the objects more accurately and robustly. For example, there are four chairs and a table in the scene of the third row. Our method can detect the ground-truth objects with almost the same bounding boxes, while other methods produce many redundant bounding boxes with even wrong category labels. Note that the results of VoteNet+3DRM are better than VoteNet while our method achieves the best results, showing the effectiveness of our method.

\begin{figure}[htb]
    \centering
    \includegraphics[width=0.41\textwidth]{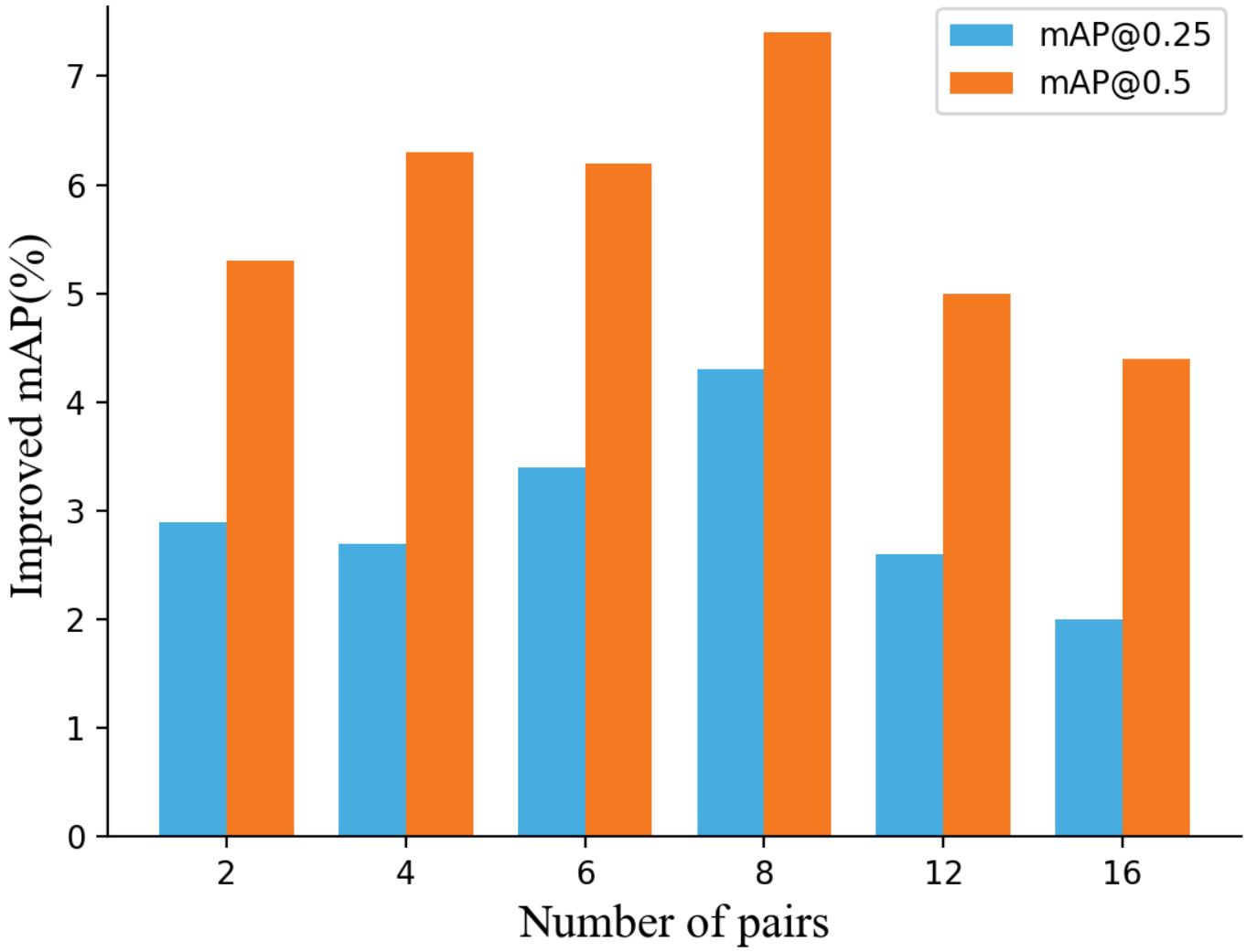}
    \caption{Improved percentage of mAP for different numbers of proposal pairs for VoteNet+ARM3D over VoteNet.}
    \label{fig:num_of_pairs}
\end{figure}

\begin{figure*}[!t]
  \centering
  \includegraphics[width=1.0\textwidth]{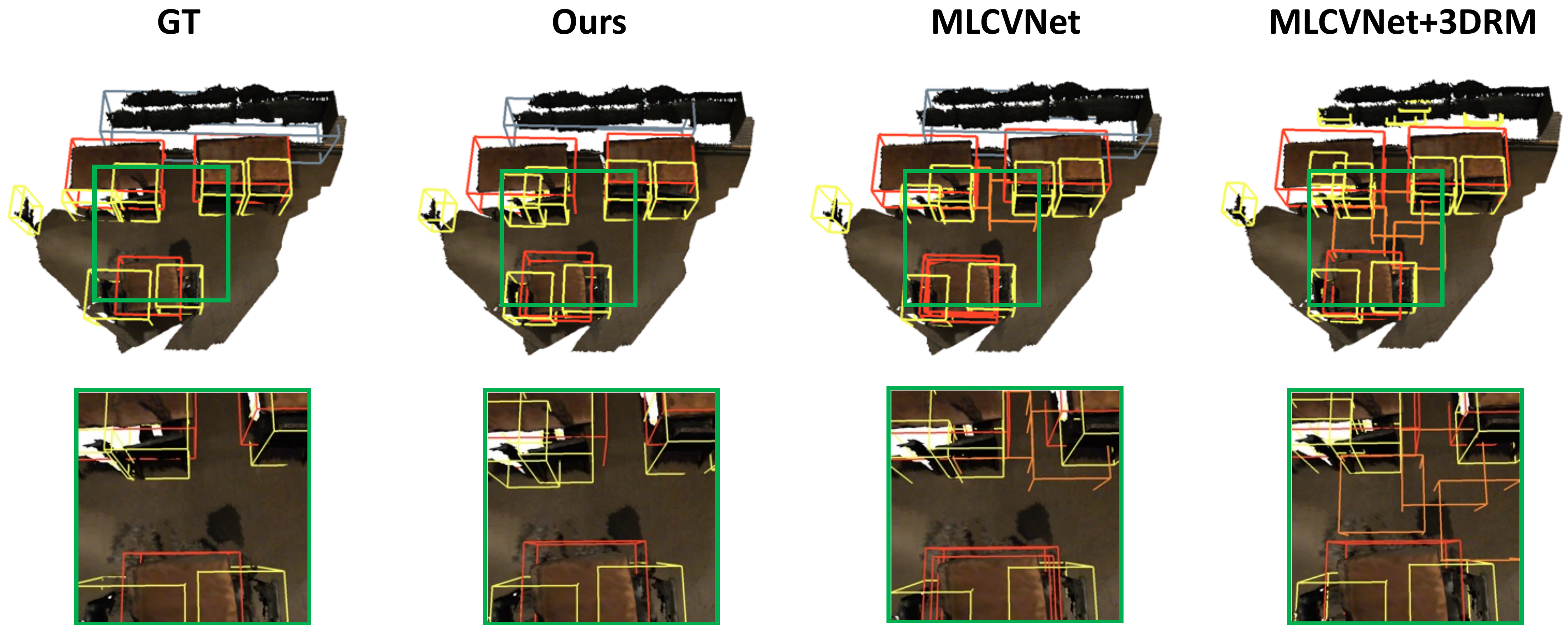}
  \caption{Qualitative comparison results of 3D object detection on ScanNetV2 val set. The detailed comparison in the second row with green rectangles demonstrates that our ARM3D enables more accurate and reasonable detection. Color is for depiction, not used for detection.}
  \label{fig:vis-scannet2}
\end{figure*}

Figure~\ref{fig:vis-sunrgbd1} displays a qualitative comparison of results from our method and other methods on the SUN RGB-D dataset. Using our method leads to better object detection with more accurate bounding boxes, while results from other methods are ambiguous or redundant. We argue that this is beneficial from our robust attention-based relation module ARM3D.

In Figure~\ref{fig:vis-scannet2}, more comparison details are displayed, and it is clear that our method achieves more robust and accurate 3D object detection. Note that the green rectangles point out the main difference between these methods, which are shown in close up in the second row. Furthro qualitative comparisons can be found in the Appendix.

The visualization of the attention examples is shown in Figure~\ref{fig:vis-attention}. On the left is the $8\times 8$ attention matrix, and on the right are weights of different proposals (dots in different colors) towards the proposal (the red dot) in the second row of the matrix. It can be seen that the proposal of a chair (the red dot) pays more attention to the sofa (the green dot) and the desk (the blue dot), corresponding to the semantic (different categories) and spatial relations (horizontal adjacency) respectively.

\begin{table*}[!t]
    \centering
    \caption{Comparison to MLCVNet and MLCVNet+3DRM$^*$ with mAP@0.25 on ScanNetV2 val set for our method with different relations. We denote MLCVNet+ARM3D as MLCVNet equipped with our ARM3D. $^*$ denotes that we retrain MLCVNet with 3DRM since 3DRM has not been applied on MLCVNet.}
    \scalebox{0.81}{
    \setlength{\tabcolsep}{0.82mm}
    \renewcommand\arraystretch{1.3}{
    \begin{tabular}{l|cccccccccccccccccc|c}
    \toprule
    \textbf{}	& wind	& bed  & cntr & sofa & tabl & showr & ofurn & sink & pic & chair & desk & curt & fridge & door & toil & bkshf & bath & cab & mAP\\  \hline
    MLCVNet & 47.0 & 88.5 & 63.9 & 87.4 & 63.5 & 65.9 & 47.9 & 59.2 & 11.9 & 90.0 & 76.1 & \textbf{56.7} & 60.9 & 56.9 & 98.3 & \textbf{56.9} & 87.2 & 42.5 & 64.5 \\
    \hline
    MLCVNet+3DRM$^*$ & 43.6 & 88.0 & 63.6 & 89.2 & 65.1 & 64.0 & 51.3 & 56.2 & 11.9 & 91.3 & 74.5 & 48.0 & 55.0 & 54.4 & 99.0 & 51.8 & 92.7 & 46.2 & 63.6 \\ 
    \hline
    MLCVNet+ARM3D(semantic) & \textbf{48.7} & 88.1 & 58.5 & \textbf{90.9} & 68.9 & 64.8 & \textbf{51.7} & 61.4 & \textbf{13.5} & 91.7 & 75.7 & 49.2 & 56.3 & \textbf{58.0} & 98.9 & 53.8 & 89.9 & 46.1 & 64.8\\

    MLCVNet+ARM3D(spatial) & 45.6 & \textbf{90.1} & 60.9 & 87.2 & 64.1 & \textbf{75.3} & 51.4 & \textbf{66.0} & 11.8 & 91.5 & 76.5 & 51.3 & \textbf{62.3} & 57.2 & \textbf{99.4} & 55.4 & 91.7 & 46.9 & 65.8\\ 
    MLCVNet+ARM3D(all) & 46.4 & 89.1 & \textbf{67.2} & 89.6 & \textbf{69.7} & 75.0 & 49.8 & 58.5 & 11.7 & \textbf{92.3} & \textbf{78.7} & 52.6 & 56.1 & 56.8 & 96.7 & 54.9 & \textbf{92.9} & \textbf{47.7} & \textbf{65.9}\\ 
    \bottomrule
    \end{tabular}}}
    \label{tab:mlcvnet-relations-025}
\end{table*}

\subsection{Ablation study}
\label{Ablation study}
\subsubsection{Effects of different components} We analyze effects of the two main components of our method including the objectness module to select proposals and the attention module. The design of the objectness module is simple but useful. It aims to select proposals with high objectness and therefore our relation module can extract reliable and robust relation context among these proposals. The attention module is to distribute different weights towards the relation context extracted from the former part since not all relation context is useful for each single proposal and some context is confusing. The objectness module and the attention module is simplified as OBM and ATM respectively in Table~\ref{tab:ablation_method}. The first row is the baseline of VoteNet+3DRM. The second row is the our method with only the objectness module and the third row is our method with only the attention module. The last row is our full method. It is noteworthy that using only OBM or ATM achieves a slight improvement. However, using both OBM and ATM, our methods obtains a larger improvement. This is attributed to our novel designs which support each other and jointly boost the performance.

\begin{table*}[!t]
  \centering
  
  \caption{Comparison of our approach VoteNet+ARM3D against VoteNet and VoteNet+3DRM with different relations on the SUN RGB-D val set with mAP@0.5.}
  \scalebox{0.95}{
  \setlength{\tabcolsep}{1.6mm}
  \renewcommand\arraystretch{1.2}{
  \begin{tabular}{l|cccccccccc|c}
  \toprule
  \textbf{}    & bathtub  & bed  & bookshelf  & chair & desk &dresser  & nightstand & sofa & table & toilet & mAP\\  \hline

  VoteNet      & 47.0  & 50.1 & 7.2 & 53.9 & 5.3 & 11.5 & 40.7 & 42.4 & 19.5 & 59.8 & 33.7  \\  \hline

  VoteNet+3DRM  & 45.4  & 51.5 & \textbf{8.5} & 55.3 & 5.5 & 16.9 & 36.8 & 48.2 & 20.5 & 62.9 & 35.1  \\    
  \hline

  VoteNet+ARM3D(semantic) & 38.7 & 51.8 & 6.4 & 57.9 & \textbf{7.1} & 	15.9 & 38.4 & \textbf{51.2} & 22.8 & \textbf{64.8} & 35.5\\   

  VoteNet+ARM3D(spatial) & 46.6 & 49.2 & 7.2 & 58.1 & 6.6 & 16.4 & 42.5 & 47.7 & 22.1 & 60.9 & 35.7\\  

  VoteNet+ARM3D(all) & \textbf{50.4} & \textbf{54.3} & 8.4 & \textbf{58.7} & 6.4 & \textbf{16.9} & \textbf{42.5} & 50.0 & \textbf{22.9} & 60.9 & \textbf{37.1}\\  
  \bottomrule
  \end{tabular}}}
  \label{tab:votenet-all-sun-05}
\end{table*}

\subsubsection{Comparison of different relations} 
The effects of different relation types we take on ScanNetV2 val dataset in terms of mAP@0.5 with regard to applying ARM3D to VoteNet are displayed in Table~\ref{tab:votenet-relations-05}. We denote VoteNet+ARM3D as our method by applying our ARM3D to VoteNet. The third row is our method with semantic relations only and the fourth row is our method with spatial relations only. The last row is our method with these two types of relations both. Using both semantic and spatial relations achieve the best performance of \textbf{7.8\%} improvement against VoteNet. Our method improves the categories of counters, showercurtains, sinks, tables and chairs by a large margin. Moreover, the detailed average precision of each category show that different categories of objects pay attention to different types of relations. For example, windows are more sensitive to spatial relations and refrigerators pay more attention to semantic relations with others, while challenging categories for detection like showercurtains and curtains need both of semantic and spatial relations for better detection. This illustrates the effectiveness and significance of both of semantic and spatial relations. A comparison of different relations of applying our method to MLCVNet on ScanNetV2 val dataset in terms of mAP@0.5 is demonstrated in Table~\ref{tab:mlcvnet-relations-025}. Similarly, our method using both semantic and spatial relations achieves the highest performance. However, MLCVNet+3DRM$^*$ reduce the performance of MLCVNet. MLCVNet is a method with three levels of rich context. The improved mAP further demonstrates the benefits and robustness of our method. The comparison on SUN RGBD val dataset for the effects of different relations to VoteNet+ARM3D in terms of mAP@0.5 can be found in Table~\ref{tab:votenet-all-sun-05}. Further comparative results are demonstrated in the Appendix.

\subsubsection{Numbers of pairs} Table~\ref{tab:ablation_pairs} shows the improved performance for differenet numbers of pairs for our method denoted as VoteNet+ARM3D on ScanNetV2. Sampling $N_k = 8$ proposal pairs for a proposal achieves the best improvement taking both mAP@0.25 and mAP@0.5 as well as computational efficiency into consideration. Further intuitive results are displayed in Figure~\ref{fig:num_of_pairs}.

\label{conclusion}
\section{Conclusions}

We propose an attention-based relation module for indoor 3D object detection on large-scale scene datasets. Using an objectness module to select raw proposals generated by backbones, we reason about the weighted relation context among themselves. Thanks to our attention module based on Transformer, we extract the most useful relation features for each proposal, which enables the network to mitigate the ambiguity and filter out those less relevant or even confusing contexts. We apply our ARM3D to two 3D object detectors on two challenging datasets for more accurate and robust detection. The consistently improved 3D object detection performance illustrates the generalization ability and effectiveness of our method.

\textbf{Future work.} Two research directions that are worth considering in future. On the one hand, it is worth trying to apply the attention-based relation module to other 3D visual tasks such as point cloud segmentation and layout arrangement. On the other hand, using a hierarchically designed relation module for reasoning about the relation contexts of sub-scenes or groups of objects is also a promising direction.

{\small
\bibliographystyle{ieee_fullname}
\bibliography{egbib}

\begin{thebibliography}{10}\itemsep=-1pt

\bibitem{atzmon2018point}
Matan Atzmon, Haggai Maron, and Yaron Lipman.
\newblock Point convolutional neural networks by extension operators.
\newblock {\em arXiv preprint arXiv:1803.10091}, 2018.

\bibitem{cadene2019murel}
Remi Cadene, Hedi Ben-Younes, Matthieu Cord, and Nicolas Thome.
\newblock Murel: Multimodal relational reasoning for visual question answering.
\newblock In {\em Proceedings of the IEEE Conference on Computer Vision and
  Pattern Recognition}, pages 1989--1998, 2019.

\bibitem{chen2019gapnet}
Can Chen, Luca~Zanotti Fragonara, and Antonios Tsourdos.
\newblock Gapnet: Graph attention based point neural network for exploiting
  local feature of point cloud.
\newblock {\em arXiv preprint arXiv:1905.08705}, 2019.

\bibitem{Chen_2020_CVPR}
Jintai Chen, Biwen Lei, Qingyu Song, Haochao Ying, Danny~Z. Chen, and Jian Wu.
\newblock A hierarchical graph network for 3d object detection on point clouds.
\newblock In {\em Proceedings of the IEEE/CVF Conference on Computer Vision and
  Pattern Recognition (CVPR)}, pages 392--401, June 2020.

\bibitem{chen2017spatial}
Xinlei Chen and Abhinav Gupta.
\newblock Spatial memory for context reasoning in object detection.
\newblock In {\em Proceedings of the IEEE International Conference on Computer
  Vision}, pages 4086--4096, 2017.

\bibitem{Chen2017Multi}
Xiaozhi Chen, Huimin Ma, Ji Wan, Bo Li, and Tian Xia.
\newblock Multi-view 3d object detection network for autonomous driving.
\newblock In {\em Proceedings of the IEEE Conference on Computer Vision and
  Pattern Recognition}, pages 1907--1915, 2017.

\bibitem{cheng2021brnet}
Bowen Cheng, Lu Sheng, Shaoshuai Shi, Ming Yang, and Dong Xu.
\newblock Back-tracing representative points for voting-based 3d object
  detection in point clouds.
\newblock In {\em Proceedings of the IEEE/CVF Conference on Computer Vision and
  Pattern Recognition}, 2021.

\bibitem{cui2020learning}
Qiongjie Cui, Huaijiang Sun, and Fei Yang.
\newblock Learning dynamic relationships for 3d human motion prediction.
\newblock In {\em Proceedings of the IEEE/CVF Conference on Computer Vision and
  Pattern Recognition}, pages 6519--6527, 2020.

\bibitem{dai2017}
Angela Dai, Angel~X Chang, Manolis Savva, Maciej Halber, Thomas Funkhouser, and
  Matthias Nie{\ss}ner.
\newblock Scannet: Richly-annotated 3d reconstructions of indoor scenes.
\newblock In {\em Proceedings of the IEEE Conference on Computer Vision and
  Pattern Recognition}, pages 5828--5839, 2017.

\bibitem{duan2019structural}
Yueqi Duan, Yu Zheng, Jiwen Lu, Jie Zhou, and Qi Tian.
\newblock Structural relational reasoning of point clouds.
\newblock In {\em Proceedings of the IEEE Conference on Computer Vision and
  Pattern Recognition}, pages 949--958, 2019.

\bibitem{Engelmann20CVPR}
Francis Engelmann, Martin Bokeloh, Alireza Fathi, Bastian Leibe, and Matthias
  Nie{\ss}ner.
\newblock 3d-mpa: Multi-proposal aggregation for 3d semantic instance
  segmentation.
\newblock In {\em Proceedings of the IEEE/CVF Conference on Computer Vision and
  Pattern Recognition}, pages 9031--9040, 2020.

\bibitem{guo2021pct}
Meng-Hao Guo, Jun-Xiong Cai, Zheng-Ning Liu, Tai-Jiang Mu, Ralph~R Martin, and
  Shi-Min Hu.
\newblock Pct: Point cloud transformer.
\newblock {\em Computational Visual Media}, 7(2):187--199, 2021.

\bibitem{hu2018relation}
Han Hu, Jiayuan Gu, Zheng Zhang, Jifeng Dai, and Yichen Wei.
\newblock Relation networks for object detection.
\newblock In {\em Proceedings of the IEEE Conference on Computer Vision and
  Pattern Recognition}, pages 3588--3597, 2018.

\bibitem{huang2018cooperative}
Siyuan Huang, Siyuan Qi, Yinxue Xiao, Yixin Zhu, Ying~Nian Wu, and Song-Chun
  Zhu.
\newblock Cooperative holistic scene understanding: Unifying 3d object, layout,
  and camera pose estimation.
\newblock In {\em Advances in Neural Information Processing Systems}, pages
  207--218, 2018.

\bibitem{huang2016structure}
Shi-Sheng Huang, Hongbo Fu, and Shi-Min Hu.
\newblock Structure guided interior scene synthesis via graph matching.
\newblock {\em Graphical Models}, 85:46--55, 2016.

\bibitem{huang2020improving}
Yifei Huang, Yusuke Sugano, and Yoichi Sato.
\newblock Improving action segmentation via graph-based temporal reasoning.
\newblock In {\em Proceedings of the IEEE/CVF Conference on Computer Vision and
  Pattern Recognition}, pages 14024--14034, 2020.

\bibitem{Krishna2016Visual}
Ranjay Krishna, Yuke Zhu, Oliver Groth, Justin Johnson, Kenji Hata, Joshua
  Kravitz, Stephanie Chen, Yannis Kalantidis, Li~Jia Li, and David~A. Shamma.
\newblock Visual genome: Connecting language and vision using crowdsourced
  dense image annotations.
\newblock {\em International Journal of Computer Vision}, 123(1):32--73, 2016.

\bibitem{ku2018joint}
Jason Ku, Melissa Mozifian, Jungwook Lee, Ali Harakeh, and Steven~L Waslander.
\newblock Joint 3d proposal generation and object detection from view
  aggregation.
\newblock In {\em 2018 IEEE/RSJ International Conference on Intelligent Robots
  and Systems (IROS)}, pages 1--8. IEEE, 2018.

\bibitem{kulkarni20193d}
Nilesh Kulkarni, Ishan Misra, Shubham Tulsiani, and Abhinav Gupta.
\newblock 3d-relnet: Joint object and relational network for 3d prediction.
\newblock In {\em Proceedings of the IEEE International Conference on Computer
  Vision}, pages 2212--2221, 2019.

\bibitem{lahoud20172d}
Jean Lahoud and Bernard Ghanem.
\newblock 2d-driven 3d object detection in rgb-d images.
\newblock In {\em Proceedings of the IEEE international conference on computer
  vision}, pages 4622--4630, 2017.

\bibitem{lan20213drm}
Yuqing Lan, Yao Duan, Yifei Shi, Hui Huang, and Kai Xu.
\newblock 3drm: Pair-wise relation module for 3d object detection.
\newblock {\em Computers \& Graphics}, 98:58--70, 2021.

\bibitem{lang2019pointpillars}
Alex~H Lang, Sourabh Vora, Holger Caesar, Lubing Zhou, Jiong Yang, and Oscar
  Beijbom.
\newblock Pointpillars: Fast encoders for object detection from point clouds.
\newblock In {\em Proceedings of the IEEE/CVF Conference on Computer Vision and
  Pattern Recognition}, pages 12697--12705, 2019.

\bibitem{li2020spatial}
Xia Li, Yibo Yang, Qijie Zhao, Tiancheng Shen, Zhouchen Lin, and Hong Liu.
\newblock Spatial pyramid based graph reasoning for semantic segmentation.
\newblock In {\em Proceedings of the IEEE/CVF Conference on Computer Vision and
  Pattern Recognition}, pages 8950--8959, 2020.

\bibitem{li2018pointcnn}
Yangyan Li, Rui Bu, Mingchao Sun, Wei Wu, Xinhan Di, and Baoquan Chen.
\newblock Pointcnn: Convolution on x-transformed points.
\newblock In {\em Advances in neural information processing systems}, pages
  820--830, 2018.

\bibitem{li2020grnet}
Ying Li, Lingfei Ma, Weikai Tan, Chen Sun, Dongpu Cao, and Jonathan Li.
\newblock Grnet: Geometric relation network for 3d object detection from point
  clouds.
\newblock {\em ISPRS Journal of Photogrammetry and Remote Sensing}, 165:43--53,
  2020.

\bibitem{Lin_2013_ICCV}
Dahua Lin, Sanja Fidler, and Raquel Urtasun.
\newblock Holistic scene understanding for 3d object detection with rgbd
  cameras.
\newblock In {\em Proceedings of the IEEE International Conference on Computer
  Vision (ICCV)}, pages 1417--1424, December 2013.

\bibitem{liu2020beyond}
Chenchen Liu, Yang Jin, Kehan Xu, Guoqiang Gong, and Yadong Mu.
\newblock Beyond short-term snippet: Video relation detection with
  spatio-temporal global context.
\newblock In {\em Proceedings of the IEEE/CVF Conference on Computer Vision and
  Pattern Recognition}, pages 10840--10849, 2020.

\bibitem{mou2019relation}
Lichao Mou, Yuansheng Hua, and Xiao~Xiang Zhu.
\newblock A relation-augmented fully convolutional network for semantic
  segmentation in aerial scenes.
\newblock In {\em Proceedings of the IEEE conference on computer vision and
  pattern recognition}, pages 12416--12425, 2019.

\bibitem{pang20163d}
Guan Pang and Ulrich Neumann.
\newblock 3d point cloud object detection with multi-view convolutional neural
  network.
\newblock In {\em 2016 23rd International Conference on Pattern Recognition
  (ICPR)}, pages 585--590. IEEE, 2016.

\bibitem{paszke2019pytorch}
Adam Paszke, Sam Gross, Francisco Massa, Adam Lerer, James Bradbury, Gregory
  Chanan, Trevor Killeen, Zeming Lin, Natalia Gimelshein, Luca Antiga, et~al.
\newblock Pytorch: An imperative style, high-performance deep learning library.
\newblock {\em Advances in neural information processing systems},
  32:8026--8037, 2019.

\bibitem{Qi_2019_ICCV}
Charles~R. Qi, Or Litany, Kaiming He, and Leonidas~J. Guibas.
\newblock Deep hough voting for 3d object detection in point clouds.
\newblock In {\em Proceedings of the IEEE/CVF International Conference on
  Computer Vision (ICCV)}, pages 9277--9286, October 2019.

\bibitem{Qi_2018_CVPR}
Charles~R Qi, Wei Liu, Chenxia Wu, Hao Su, and Leonidas~J Guibas.
\newblock Frustum pointnets for 3d object detection from rgb-d data.
\newblock In {\em Proceedings of the IEEE conference on computer vision and
  pattern recognition}, pages 918--927, 2018.

\bibitem{Charles2017PointNet}
Charles~R Qi, Hao Su, Kaichun Mo, and Leonidas~J Guibas.
\newblock Pointnet: Deep learning on point sets for 3d classification and
  segmentation.
\newblock In {\em Proceedings of the IEEE conference on computer vision and
  pattern recognition}, pages 652--660, 2017.

\bibitem{qi2017pointnet++}
Charles~Ruizhongtai Qi, Li Yi, Hao Su, and Leonidas~J Guibas.
\newblock Pointnet++: Deep hierarchical feature learning on point sets in a
  metric space.
\newblock {\em Advances in neural information processing systems},
  30:5099--5108, 2017.

\bibitem{qi20173d}
Xiaojuan Qi, Renjie Liao, Jiaya Jia, Sanja Fidler, and Raquel Urtasun.
\newblock 3d graph neural networks for rgbd semantic segmentation.
\newblock In {\em Proceedings of the IEEE International Conference on Computer
  Vision}, pages 5199--5208, 2017.

\bibitem{ren2015faster}
Shaoqing Ren, Kaiming He, Ross Girshick, and Jian Sun.
\newblock Faster r-cnn: Towards real-time object detection with region proposal
  networks.
\newblock {\em Advances in neural information processing systems}, 28:91--99,
  2015.

\bibitem{santoro2017simple}
Adam Santoro, David Raposo, David~G Barrett, Mateusz Malinowski, Razvan
  Pascanu, Peter Battaglia, and Timothy Lillicrap.
\newblock A simple neural network module for relational reasoning.
\newblock In {\em Advances in neural information processing systems}, pages
  4967--4976, 2017.

\bibitem{shi2019pointrcnn}
Shaoshuai Shi, Xiaogang Wang, and Hongsheng Li.
\newblock Pointrcnn: 3d object proposal generation and detection from point
  cloud.
\newblock In {\em Proceedings of the IEEE Conference on Computer Vision and
  Pattern Recognition}, pages 770--779, 2019.

\bibitem{shi2020points}
Shaoshuai Shi, Zhe Wang, Jianping Shi, Xiaogang Wang, and Hongsheng Li.
\newblock From points to parts: 3d object detection from point cloud with
  part-aware and part-aggregation network.
\newblock {\em IEEE transactions on pattern analysis and machine intelligence},
  2020.

\bibitem{shi2019hierarchy}
Yifei Shi, Angel~X Chang, Zhelun Wu, Manolis Savva, and Kai Xu.
\newblock Hierarchy denoising recursive autoencoders for 3d scene layout
  prediction.
\newblock In {\em Proceedings of the IEEE Conference on Computer Vision and
  Pattern Recognition}, pages 1771--1780, 2019.

\bibitem{shi2016data}
Yifei Shi, Pinxin Long, Kai Xu, Hui Huang, and Yueshan Xiong.
\newblock Data-driven contextual modeling for 3d scene understanding.
\newblock {\em Computers \& Graphics}, 55:55--67, 2016.

\bibitem{song2017web3d}
Peihua Song, Youyi Zheng, and Jinyuan Jia.
\newblock Web3d learning platform of furniture layout based on case-based
  reasoning and distance field.
\newblock In {\em International Conference on Technologies for E-Learning and
  Digital Entertainment}, pages 235--250. Springer, 2017.

\bibitem{song2015}
Shuran Song, Samuel~P Lichtenberg, and Jianxiong Xiao.
\newblock Sun rgb-d: A rgb-d scene understanding benchmark suite.
\newblock In {\em Proceedings of the IEEE conference on computer vision and
  pattern recognition}, pages 567--576, 2015.

\bibitem{sun2020dagc}
Qi Sun, Hongyan Liu, Jun He, Zhaoxin Fan, and Xiaoyong Du.
\newblock Dagc: Employing dual attention and graph convolution for point cloud
  based place recognition.
\newblock In {\em Proceedings of the 2020 International Conference on
  Multimedia Retrieval}, pages 224--232, 2020.

\bibitem{sung2018learning}
Flood Sung, Yongxin Yang, Li Zhang, Tao Xiang, Philip~HS Torr, and Timothy~M
  Hospedales.
\newblock Learning to compare: Relation network for few-shot learning.
\newblock In {\em Proceedings of the IEEE Conference on Computer Vision and
  Pattern Recognition}, pages 1199--1208, 2018.

\bibitem{wang2019graph}
Lei Wang, Yuchun Huang, Yaolin Hou, Shenman Zhang, and Jie Shan.
\newblock Graph attention convolution for point cloud semantic segmentation.
\newblock In {\em Proceedings of the IEEE/CVF Conference on Computer Vision and
  Pattern Recognition}, pages 10296--10305, 2019.

\bibitem{wang2017cnn}
Peng-Shuai Wang, Yang Liu, Yu-Xiao Guo, Chun-Yu Sun, and Xin Tong.
\newblock O-cnn: Octree-based convolutional neural networks for 3d shape
  analysis.
\newblock {\em ACM Transactions On Graphics (TOG)}, 36(4):1--11, 2017.

\bibitem{wang2019exploring}
Wenbin Wang, Ruiping Wang, Shiguang Shan, and Xilin Chen.
\newblock Exploring context and visual pattern of relationship for scene graph
  generation.
\newblock In {\em Proceedings of the IEEE Conference on Computer Vision and
  Pattern Recognition}, pages 8188--8197, 2019.

\bibitem{wang2019deep}
Yue Wang and Justin~M Solomon.
\newblock Deep closest point: Learning representations for point cloud
  registration.
\newblock In {\em Proceedings of the IEEE/CVF International Conference on
  Computer Vision}, pages 3523--3532, 2019.

\bibitem{wen2021airborne}
Congcong Wen, Xiang Li, Xiaojing Yao, Ling Peng, and Tianhe Chi.
\newblock Airborne lidar point cloud classification with global-local graph
  attention convolution neural network.
\newblock {\em ISPRS Journal of Photogrammetry and Remote Sensing},
  173:181--194, 2021.

\bibitem{wen2020point}
Xin Wen, Tianyang Li, Zhizhong Han, and Yu-Shen Liu.
\newblock Point cloud completion by skip-attention network with hierarchical
  folding.
\newblock In {\em Proceedings of the IEEE/CVF Conference on Computer Vision and
  Pattern Recognition}, pages 1939--1948, 2020.

\bibitem{wu2019pointconv}
Wenxuan Wu, Zhongang Qi, and Li Fuxin.
\newblock Pointconv: Deep convolutional networks on 3d point clouds.
\newblock In {\em Proceedings of the IEEE/CVF Conference on Computer Vision and
  Pattern Recognition}, pages 9621--9630, 2019.

\bibitem{Xie_2020_CVPR}
Qian Xie, Yu-Kun Lai, Jing Wu, Zhoutao Wang, Yiming Zhang, Kai Xu, and Jun
  Wang.
\newblock Mlcvnet: Multi-level context votenet for 3d object detection.
\newblock In {\em Proceedings of the IEEE/CVF Conference on Computer Vision and
  Pattern Recognition (CVPR)}, pages 10447--10456, June 2020.

\bibitem{Xu_2019_CVPR}
Hang Xu, Chenhan Jiang, Xiaodan Liang, and Zhenguo Li.
\newblock Spatial-aware graph relation network for large-scale object
  detection.
\newblock In {\em Proceedings of the IEEE Conference on Computer Vision and
  Pattern Recognition}, pages 9298--9307, 2019.

\bibitem{xu2012fit}
Kai Xu, Hao Zhang, Daniel Cohen-Or, and Baoquan Chen.
\newblock Fit and diverse: Set evolution for inspiring 3d shape galleries.
\newblock {\em ACM Transactions on Graphics (TOG)}, 31(4):1--10, 2012.

\bibitem{xu2011photo}
Kai Xu, Hanlin Zheng, Hao Zhang, Daniel Cohen-Or, Ligang Liu, and Yueshan
  Xiong.
\newblock Photo-inspired model-driven 3d object modeling.
\newblock {\em ACM Transactions on Graphics (TOG)}, 30(4):1--10, 2011.

\bibitem{yan2018second}
Yan Yan, Yuxing Mao, and Bo Li.
\newblock Second: Sparsely embedded convolutional detection.
\newblock {\em Sensors}, 18(10):3337, 2018.

\bibitem{yang20203dssd}
Zetong Yang, Yanan Sun, Shu Liu, and Jiaya Jia.
\newblock 3dssd: Point-based 3d single stage object detector.
\newblock In {\em Proceedings of the IEEE/CVF conference on computer vision and
  pattern recognition}, pages 11040--11048, 2020.

\bibitem{yew20183dfeat}
Zi~Jian Yew and Gim~Hee Lee.
\newblock 3dfeat-net: Weakly supervised local 3d features for point cloud
  registration.
\newblock In {\em Proceedings of the European Conference on Computer Vision
  (ECCV)}, pages 607--623, 2018.

\bibitem{yi2019gspn}
Li Yi, Wang Zhao, He Wang, Minhyuk Sung, and Leonidas~J Guibas.
\newblock Gspn: Generative shape proposal network for 3d instance segmentation
  in point cloud.
\newblock In {\em Proceedings of the IEEE/CVF Conference on Computer Vision and
  Pattern Recognition}, pages 3947--3956, 2019.

\bibitem{zhang2019pcan}
Wenxiao Zhang and Chunxia Xiao.
\newblock Pcan: 3d attention map learning using contextual information for
  point cloud based retrieval.
\newblock In {\em Proceedings of the IEEE/CVF Conference on Computer Vision and
  Pattern Recognition}, pages 12436--12445, 2019.

\bibitem{deepcontext}
Yinda Zhang, Mingru Bai, Pushmeet Kohli, Shahram Izadi, and Jianxiong Xiao.
\newblock Deepcontext: Context-encoding neural pathways for 3d holistic scene
  understanding.
\newblock In {\em 2017 IEEE International Conference on Computer Vision
  (ICCV)}, pages 1201--1210, 10 2017.

\bibitem{H3DNet}
Zaiwei Zhang, Bo Sun, Haitao Yang, and Qixing Huang.
\newblock H3dnet: 3d object detection using hybrid geometric primitives.
\newblock In Andrea Vedaldi, Horst Bischof, Thomas Brox, and Jan-Michael Frahm,
  editors, {\em Proceedings of the European Conference on Computer Vision
  (ECCV)}, pages 311--329, Cham, 2020. Springer International Publishing.

\bibitem{zhao2020point}
Hengshuang Zhao, Li Jiang, Jiaya Jia, Philip Torr, and Vladlen Koltun.
\newblock Point transformer.
\newblock {\em arXiv preprint arXiv:2012.09164}, 2020.

\bibitem{zhao2018triangle}
Yawei Zhao, Kai Xu, En Zhu, Xinwang Liu, Xinzhong Zhu, and Jianping Yin.
\newblock Triangle lasso for simultaneous clustering and optimization in graph
  datasets.
\newblock {\em IEEE Transactions on Knowledge and Data Engineering},
  31(8):1610--1623, 2018.

\end{thebibliography}
}

\end{document}